\documentclass[12pt]{article}

\newtheorem{remark}{Remark}

\usepackage[colorlinks,
            linkcolor=blue,
            anchorcolor=blue,
            citecolor=blue]{hyperref}
\usepackage{graphicx}
\usepackage{booktabs}
\usepackage{threeparttable}
\usepackage{subfigure}
\usepackage{footnote}
\usepackage{tabu}
\usepackage{tabularx}

\usepackage{changes}
\definechangesauthor[name={Q.L.}, color=orange]{Q.}

\usepackage{subfigure}
\usepackage[ruled]{algorithm2e}
\usepackage{float}

\usepackage[numbers,sort&compress]{natbib}

\usepackage{geometry}
\def\bq{\begin{equation}}
\def\eq{\end{equation}}
\def\bea{\begin{eqnarray}}
\def\eea{\end{eqnarray}}

\title{Clustering-Based Representation Learning through Output Translation and Its Application to Remote--Sensing Images}
\author{Qinglin Li$^{1,2}$, Bin Li$^{1,2}$, Jonathan M. Garibaldi$^3$ and Guoping Qiu$^{1,2,4,5,*}$\\
$^{1}$ College of Electronic and Information Engineering, Shenzhen University,\\ Shenzhen 518052, China; qlilx@szu.edu.cn (Q.L.); libin@szu.edu.cn (B.L.)\\
$^{2}$  Guangdong Key Lab for Intelligent Information Processing, \\ Shenzhen University, Shenzhen 518052, China.\\
$^{3}$ School of Computer Science, The University of Nottingham,\\ Nottingham NG8 1BB, UK; jon.garibaldi@nottingham.ac.uk\\
$^{4}$ Shenzhen Institute of AI and Robotics for Society, \\Shenzhen 518172, China\\
$^{5}$ Pengcheng Laboratory, Shenzhen 518055, China\\
$^{*}$Correspondence: qiu@szu.edu.cn
}

\date{}

\begin{document}
\maketitle

\begin{abstract}

In supervised deep learning, learning good representations for remote--sensing images (RSI) relies on manual annotations. However, in the area of remote sensing, it is hard to obtain huge amounts of labeled data. Recently, self--supervised learning shows its outstanding capability to learn representations of images, especially the methods of instance discrimination. Comparing methods of instance discrimination, clustering--based methods not only view the transformations of the same image as ``positive" samples but also similar images. In this paper, we propose a new clustering--based method for representation learning. We first introduce a quantity to measure representations' discriminativeness and from which we show that even distribution requires the most discriminative representations. This provides a theoretical insight into why evenly distributing the images works well. We notice that only the even distributions that preserve representations' neighborhood relations are desirable. Therefore, we develop an algorithm that translates the outputs of a neural network to achieve the goal of evenly distributing the samples while preserving outputs' neighborhood relations. Extensive experiments have demonstrated that our method can learn representations that are as good as or better than the state of the art approaches, and that our method performs computationally efficiently and robustly on various RSI datasets. The paper has been published on Remote Sensing \url{https://doi.org/10.3390/rs14143361}. 

\end{abstract}




\section{Introduction}

The self--supervised visual--representation learning paradigm, where ``supervision'' is created from making use of the intrinsic information contained in the data itself and based on prior knowledge about the world, provides us with an unsupervised way to learn representations of RSI. Recently, methods of instance discrimination~\cite{journals/corr/abs-1807-03748, conf/cvpr/YeZYC19,journals/corr/abs-1905-09272,conf/eccv/TianKI20,conf/cvpr/He0WXG20,journals/corr/abs-2002-05709} have achieved remarkable progress in unsupervised representation learning. These methods view every sample and its transformations as a ``class'', while, in clustering--based methods~\cite{conf/eccv/CaronBJD18,conf/iclr/AsanoRV20a}, samples grouped to the same cluster are considered as a ``class''. In fact, data augmentations are often used in clustering--based methods, and, thus, transformations of the same images are also viewed as the same ``class''. However, so far, methods of instance discrimination overall have better performances than clustering--based methods, which seems to imply that clustering--based methods could be further improved.

Degeneracy is a problem that nearly all clustering--related methods need to solve. The usual ways to tackle it is by setting constraints on the clustering process. Whether it is based on maximizing entropy~\cite{jconf/iccv/JiVH19,conf/eccv/GansbekeVGPG20} or adding regularization \cite{conf/iccv/DizajiHDCH17} , all aim to distribute the samples evenly to the clusters. However, even distribution is not a necessary condition for preventing degeneracy; there is a lack of a clear theoretical interpretation for why even distribution works well in these clustering--related methods. On the other hand, clustering should be based on the neighborhood relations of the samples or that of their representations. To avoid degeneracy, the normal clustering process (that clusters should be formed based on the relations amongst the samples) would have to be artificially introduced. It is unclear if approaches used by previous methods to achieve even distribution can completely preserve the neighborhood relations of samples or that of their representations. By closely observing the Sinkhorn--Knopp algorithm used by~\cite{conf/iclr/AsanoRV20a}, it is easy to see that the relations between neighboring samples or the distances between the outputs have been changed by the intervention to achieve even distribution.

In this paper, we first introduce a quantity that can be used to measure how discriminative representations are under a generic clustering setting where the clusters are formed based {on } the neighborhood relations or similarity of an images' representations. From this quantity, we show that evenly distributing the images into clusters requires the visual representations to be most discriminative. This provides an insight that clearly explains why evenly distributing the samples into clusters worked well in previous related work~\cite{conf/eccv/GansbekeVGPG20,conf/iclr/AsanoRV20a}. However, we are only interested in the even distributions that preserve the model's smoothness, i.e., preserve the neighborhood relations or similarity of the images' representations, and, thus, we develop an algorithm\footnote{The code 
is avaible at \url{https://github.com/qlilx/OTL}} that translates the outputs of a convolutional neural network to achieve the goal of evenly distributing the images while, at the same time, preserving the current model's smoothness. In summary, we make the following contributions:
\begin{itemize}
    \item A quantity that can be used to measure the discriminativeness of visual representations is introduced and a clear theoretical explanation for evenly distributing images into clusters is given from the viewpoints of this quantity. 
    \item A new efficient and robust clustering--based representation learning algorithm, through translating the outputs of a convolutional neural network, is proposed.   
    \item Our method performs as well as or better than the state--of--the--art approaches and works very well on a variety of RSI datasets. 
\end{itemize}

\section{Related Works}

Generally, most self--supervised methods can be categorized into four categories: clustering--based methods, contrastive methods, pretext tasks and generative models.

\textbf{Clustering--based} methods learn visual representations by integrating network optimization and clustering. Clustering is a classical approach to unsupervised machine learning~\cite{DBLP:books/crc/aggarwal2013}. K--means, a standard clustering algorithm, has been applied to DeepCluster~\cite{conf/eccv/CaronBJD18} for grouping the features extracted from a deep neural network. Subsequently,  the cluster assignments  were utilized as supervision to update the weights of the network. The method, Anchor {Neighborhood }  Discovery (AND)~\cite{conf/icml/HuangDGZ19} exploits a class--consistent  {neighborhood }for unsupervised learning, which combines the advantages of both sample--specificity learning and clustering while overcoming their disadvantages. Maximizing the information between the indices of inputs and labels, self--labeling~\cite{conf/iclr/AsanoRV20a} not only avoids the degenerated problem but also creates pseudo--labels for training a deep neural network by a standard cross--entropy loss. Other earlier works of combining clustering and deep learning can be found in~\cite{conf/icml/XieGF16,DBLP:conf/icml/YangFSH17,conf/nips/LiaoSZU16,conf/nips/DosovitskiySRB14,conf/cvpr/YangPB16}.

\textbf{The contrastive method}~\cite{journals/corr/abs-1807-03748, conf/cvpr/YeZYC19,journals/corr/abs-1905-09272,conf/eccv/TianKI20,conf/cvpr/He0WXG20,journals/corr/abs-2002-05709}
has recently shown remarkable performances on visual representation learning. The contrastive method is a method of performing instance discrimination, which only considers the transformations of the same image as the ``positive'' samples, while others are the ``negative'' samples in the contrastive loss. Recently, some methods combining clustering methods and contrastive methods have appeared. In~\cite{Dwibedi2021WithAL}, the nearest neighbors are also considered as the ``positive'' samples in the contrastive loss, and in~\cite{Li2021ContrastiveC}, both instance--level and cluster--level contrastive losses are introduced. In the past year, contrastive methods have been extensively applied to area of remote sensing \cite{9397864,rs13214255,9522871,li2022global}.

\textbf{A pretext task} uses hand--crafted ``supervision'' to replace manual labels in supervised learning. These ``supervision'' pretext tasks, are devised from exploiting the intrinsic information of the unlabeled data. These pretext tasks include predicting context~\cite{conf/iccv/DoerschGE15}, solving jigsaw puzzles~\cite{conf/eccv/NorooziF16,conf/wacv/KimCYK18}, image rotation and colorization~\cite{conf/iclr/GidarisSK18,conf/cvpr/KolesnikovZB19,conf/eccv/ZhangIE16,conf/eccv/LarssonMS16}, spatio--temporal consistence~\cite{conf/iccv/WangG15}, and so on. The applications of this method to RSI can be found in \cite{akiva2021self,ayush2021geography,rs12111868}, which use domain knowledge and temporal prediction to supervise the training.

\textbf{Generative models.} The latent distributions in generative models could be visual representations, due to  { the fact that} they can capture the distributions of input data. These generative models mainly include Boltzmann machines~\cite{HintonETAL:06,conf/icml/LeeGRN09,conf/cvpr/TangSH12}, autoencoders~\cite{JMLR:v11:vincent10a} and generative adversarial networks (GAN)~\cite{conf/cvpr/RenL18,conf/cvpr/JenniF18,conf/nips/XieDDHN17,conf/nips/DonahueS19}. In \cite{rs14102425}, an improved generative adversarial network was applied to the super--resolution processing of RSI.

\begin{figure}[H]
  \centering
  \includegraphics[width=3.5in]{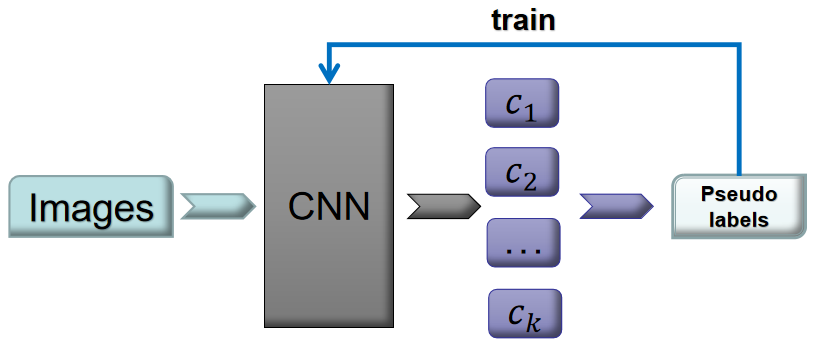}\\
  \caption{Schematics of representation learning based on clustering: input images are clustered into $k$ clusters according to representations (or outputs) extracted from the CNN; and then the identities of clusters are utilized as pseudo--labels to train the CNN.}
  \label{cl_rp}
\end{figure}

\section{Method}\label{method}

 {The goal of this paper is to learn good visual representations for images through clustering them; that is, clustering is not the target but a means to learn visual representation. This section will introduce two essential aspects of good visual representations and, based on them, design our algorithm and training method.}

\subsection{Discriminativeness of Representations}
\label{discrmination}

In the representation feature space, the feature vectors should reveal as much as possible the difference in nonidentical samples; that is, good representations should be discriminative. To compare which representations are more discriminative, we need to be able to measure the discriminativeness of representations. In this section, based on the generic clustering model as shown in Figure \ref{cl_rp}, we try to find a criterion that can be used to compare the discriminativeness of representations.

Consider a general clustering that groups $N$ images, based on their representations in feature space $F$, into $k$ clusters by certain methods, for example, k--means. 
Samples in the same cluster will be assigned the same pseudo--labels that will be applied to standard cross entropy loss to update the network. Due to samples in the same cluster having closer representations and being indiscriminately treated, we could legitimately regard the representations of samples from the same cluster as indistinguishable.  If the $N$ representations in space $F$ are very discriminative, there will be 
few indistinguishable representations.

Therefore, the quantity that can be used to compare the discriminativeness of representations should be able to indicate the representations' degree of distinguishableness. When we say ``distinguishable'',   {when at least two representations are needed to be compared}. Consequently, in this paper, we use the number of distinguishable pairwise comparisons of representations as the quantity to measure the discriminativeness of representations. 

In different representation feature spaces, the clustering results will be different. Let $n_i(F)$ be the number of samples assigned into cluster $i$ in the feature space $F$, and, {thus, $N = n_1(F) + n_2(F) + \cdots + n_k(F)$, where $N$ is the total number of samples}. The total number of pairwise comparisons of $N$ representations is $N(N-1)/2$, in which the number of indistinguishable pairwise comparisons is 
\bq\label{ns}
N_{\textrm{ind}}(F)= \sum_i^k
\left(
\begin{array}{c}
     n_i \\
     2 
\end{array}
\right)
\eq
and the number of distinguishable pairwise comparisons is 
\bq\label{nd}
N_{\textrm{dis}}(F) = \sum_{i=1}^{k-1}\left(n_i(F)\sum_{j=i+1}^kn_j(F)\right).
\eq

Consider representations in two different feature spaces $F_1$ and $F_2$, and perform clustering in these two spaces. 
If 
\bq\label{f12}
N_{\textrm{dis}} (F_1) > N_{\textrm{dis}} (F_2), 
\eq
then the representations in $F_1$ ought to be more discriminative than those in $F_2$ because there are more distinguishable representations in $F_1$. Therefore, the quantity $N_{\textrm{dis}}(F)$ can be used to measure the discriminativeness of the $N$ samples' representations in a feature space $F$: the larger is $N_{\textrm{dis}}(F)$, the more discriminative are the representations. Several simple toy examples are given in Appendix {\ref{sim-exam}} to elaborate the rationality of this quantification.

From the view points of $N_{\textrm{dis}}(F)$, the degeneracy problem in clustering methods is caused by small $N_{\textrm{dis}}(F)$. For example, the extreme case that $N_{\textrm{dis}}(F)=0$ means that the representations are totally indistinguishable, which corresponds to the case where all samples are assigned into the same cluster. Therefore, increasing $N_{\textrm{dis}}(F)$ can prevent the representations from degenerating. However, the question is, when will $N_{\textrm{dis}}(F)$ reach the maximum?

Note that $N_{\textrm{ind}} + N_{\textrm{dis}} = N(N-1)/2 $. Thus, the maximum $N_{\textrm{dis}}(F)$ corresponds to the minimal $N_{\textrm{ind}}(F)$ due to $N$ and $k$ being constants. {Equation (\ref{ns}) simplifies to }
\bea\label{miniv}
N_{\textrm{ind}} =\frac{1}{2} \left(\sum^k_in^2_i(F)-N\right).
\eea
Again, noting $N$ is a constant, $N_{\textrm{ind}}$ achieves the minimal value when $\sum^k_in^2_i(F)$ is the minimal. According to the inequality of arithmetic and geometric means (AM--GM inequality) \cite{cauchy_2009} and noting $n_i(F) \geq0$,   $\sum^k_in^2_i(F)$ takes the minimal value if
\bq\label{evcondn}
 n_i(F)=\bar n(F)=N/k,~~\forall i \in \{1,2\cdots,k\}
\eq
which is also the condition for $N_{\textrm{ind}}$ to reach the maximum. For the cases when $\bar n = N/k $ is not an integer, a rigorous proof is given in Appendix {\ref{appd}} that $N_{\textrm{ind}}$ reaches the largest value when the standard deviation of the distribution of $n_i(F)$ is the minimal, i.e., {the N samples are distributed as uniformly (evenly) as possible among the $k$ clusters.}

\begin{remark}
\label{md}
 If $N$ nonidentical images are most evenly distributed into $k$ clusters according to their representations in feature space $F$, then $N_{\textrm{ind}}(F)$ must be the maximum.
\end{remark}


{The fact stated in Remark \ref{md} } gives an explicit explanation for why even distribution works well in previous clustering--related methods, e.g.,~\cite{conf/iclr/AsanoRV20a}. From the view point of  {Remark~\ref{md}} : an even distribution corresponds to the most discriminative representations, which can be regarded as the extreme opposite of the degenerated cases. However, for previous works in the literature, the even distribution is simply used as a tool to prevent the model from collapsing. In contrast, in this paper, we provide a fundamental insight that has enabled us to give a clear interpretation for even distribution and use it to develop new techniques to improve the discriminativeness of representations.

Intuitively, ``the most discriminative representations give the most uniform distribution'' is correct only for special cases. It is important to emphasize that, in this paper, we measure the discriminativeness of representations by $N_{\textrm{ind}}(F)$, which gives how many pairs of representations are distinguishable. Again, the rationality of this quantification can be seen from simple examples in Appendix {\ref{sim-exam}}.

 For the given $N$ and $k$, when $n_i(F)$ is the most uniform distribution,
 the assignment of the input images to the $k$ clusters has $N_e$ possible combinations
\bea
N_e = \frac{1}{k!}
\left(
  \begin{array}{c}
    N \\
    n_1(F) \\
  \end{array}
 \right)
\prod^k_{i=2}
\left(
\begin{array}{cc}
    N - \sum^{i-1}_{j=1}n_{j}(F) \\
    n_i(F)
\end{array}
\right).
\eea
This means that the solution to maximal $N_{\textrm{ind}}(F)$ is not unique. Therefore, the question now is how we shall approach maximum $N_{\textrm{ind}}(F)$ such that it is good for representation learning. 

\subsection{General--Purpose Prior: Smoothness}\label{smoothness}

Good representation should be discriminative. However, at the end of Section \ref{discrmination}, we demonstrated that there are many ways to reach evenly distributed pseudo--labels that can be used to improve the discriminativeness of representations. We are only interested in the ones that are good for representation learning. Recall that the basic task of representation learning is to make similar samples to have close representations while different samples have distinct representations. In other words, representation learning is the pursuit of a smooth learning model tht satisfies 
\bq\label{sm}
f(x_1) \approx f(x_2)~~\textrm{iff}~~x_1 \approx x_2,
\eq
where $f$ is the mapping function of the model to be learnt. Smoothness is a basic and general requirement of a learning model \cite{6472238}. Therefore, only the pseudo--labels that are good for improving the smoothness of the model are desirable. Specifically, this requires that similar samples are more likely to be assigned the same pseudo--labels. 

Some model architectures have a certain inherent smoothness. For example, randomly initialized AlexNet possesses some degree of smoothness \cite{conf/eccv/NorooziF16}.  Therefore, clustering based on the outputs or internal features extracted by the model is likely to group close samples together, thus assigning them with the same pseudo--labels. Applying these pseudo--labels in the standard cross--entropy loss can enhance the smoothness of the model. Besides this, making the representations of transformations of the same image close to each other can also improve the smoothness of the model. Therefore, without any manual annotation, assigning pseudo--labels based on the neighborhood relations of the outputs or the internal representations of the network is a reasonable strategy. This provides us with an important constraint on approaching the even distribution of images in clusters, that is, the {neighborhood}  relations of the outputs or the internal representations of the network must be protected. Recall that evenly assigning the pseudo--labels is necessary because it can not only prevent the model from collapsing, but also improve the discriminativeness of~representations.

\subsection{Algorithm}\label{algorithm}

This section proposes an effective algorithm that can produce evenly distributed pseudo--labels according to the outputs of the learning model while keeping the smoothness of the current model unaltered. We then use the pseudo--labels and a label--consistent--training (LCT) technique to improve the model's smoothness.

\textbf{Model architecture.} We consider a general convolutional neural network (CNN) model, as shown in Figure \ref{cl_rp}, for representation learning. One or several fully connected layers are attached at the end of the CNNs, which will output a $k$--dimensional vector indicating the assignment of the input samples to the $k$ clusters. Specifically, the $s$th input images $I(s)$ is mapped to $X(s)$ by the convolutional block $\phi$, i.e., $X(s) =\phi(I(s))$.  By one or several fully connected layers, the feature $X(s)$ in $D$--dimensional space is then mapped to a $k$--dimensional space: $R^D\rightarrow R^k$
\bq\label{out}
O(s)=\big(o_1(s), o_2(s), \cdots, o_k(s)\big),
\eq
where $O(s)$ is the output of $I(s)$ {without softmax operation (in this paper, all ``output`` refers to the final layer of network but the softmax activation on the final layer). } {The output dimension is set to be the same as the cluster number, such that images could be assigned into $k$ clusters based on the rule:} the input $I(s)$ is assigned to cluster $c_i$ if and only if $o_i(s)$ is the maximal component:
\bq\label{ci}
I(s)\in c_i~~~\textrm{iff}~~~o_i(s) \geq o_j(s), ~~~\forall j\neq i, \;\;i,j = 1, 2,\cdots,k.
\eq
 Clearly, (\ref{ci}) is a winner--takes--all competition extensively used in unsupervised competitive learning \cite{conf/icnn/DeSieno88}. The largest output wins the competition and the rest lose out. 

\begin{figure}[H]
  \centering
  \includegraphics[width=4.5in]{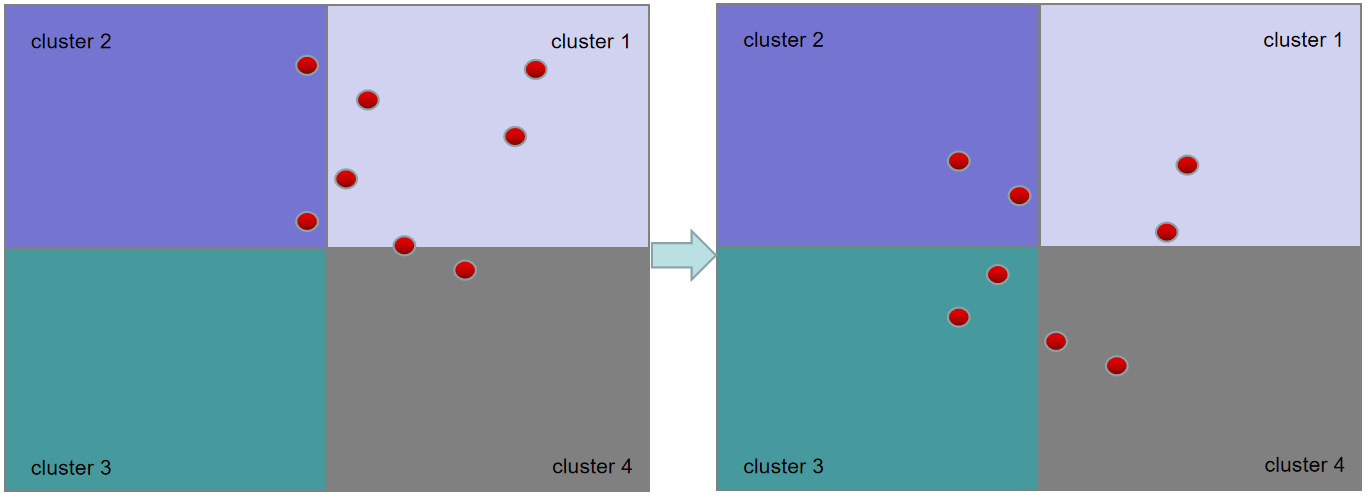}\\
  \caption{{Left: most points are located in the first quadrant and thus are assigned into the cluster 1, while there are no points located in the third quadrant; right: Translating all points as a whole not only can prevent their relative distances, but also make them evenly distributed into the four~quadrants.  }
  }\label{move-out}
\end{figure}

\textbf{Output translation.} In most cases, if we perform clustering as (\ref{ci}) based on outputs (\ref{out}), most input samples will be assigned into a few clusters. {Consider a toy model: assigning eight points in a two-dimensional plane to four clusters based on the quadrant the points are located in. Usually, based on the original location of points, they would not be evenly distributed into the four clusters; for example, the first graph of Figure \ref{move-out} (one red point represents a point, the boundaries in the diagram are the $x$ and $y$ axes). If the points need to be evenly distributed into the four clusters based on the same rule, the values of their coordinates have to be changed and the question is how to change the values of them. If there is a constraint that their relative distances should be kept (the smoothness requirement), translating them as whole is the most straightforward approach to carry that out, as shown in the second graph of Figure \ref{move-out}. Considering $N$ points in $k$ dimensions and changing the clustering rule to Equation (\ref{ci}), the model then becomes what this paper needs to study and, similarly, the even distribution of pseudo--labels could be reached by translating the outputs in $k$ dimensions. }

Therefore, the core task of our algorithm is that of finding the translation $T$
\bea\label{trans}
O'(s) = O(s) - T
\eea
such that cluster assignment according to (\ref{ci}) using the new outputs $O'(s)$ gives evenly distributed pseudo--labels. Obviously, $T$ is a $k$--dimensional vector: $T=(t_1, t_2, \cdots, t_k)$.
To find $t_1, t_2, \cdots, t_k$, we borrow the basic idea from a competitive learning or self--organising learning algorithm called frequency sensitive competitive learning (FSCL), widely used in the design of optimal vector quantisation \cite{48637,journals/nn/AhaltKCM90}. If one component (or a direction) ``wins'' too much in the competition  (\ref{ci}), we move all output vectors away from this direction; if one direction ``loses'' the competition too often, we move all output vectors towards this direction. To perform this, we need to construct each direction's wining frequency indicator in features space $F$. We count the number of samples assigned into every cluster $c_i$ according to (\ref{ci}), and set the number as its wining frequency indicator. For convenience, we subtract the mean  $\bar n(F) =N/k$ from each cluster to obtain the frequency indicator as 
\bea\label{indicator}
C= \left(
\begin{array}{c}
n_1(F) - \bar n(F) \\
n_2(F) - \bar n(F) \\
\cdots \\
n_k(F) - \bar n(F)
\end{array} \right)
=
\left(
\begin{array}{c}
\hat n_1(F) \\
\hat n_2(F) \\
\cdots \\
\hat n_k(F)
\end{array} \right)
\eea
where $\hat n_i(F) = n_i(F) - \bar n(F)$. $C$ can be used as all directions' winning frequency indicator, e.g., the winning frequency of the first direction is $\hat{n}_1(F)$. Note that some $\hat n_i(F)$ are positive while others are non--positive. The vector of translation should be proportional to this indicator $T\propto C$. 
The frequency indicator determines the direction of translation. The magnitude of the translation should be close to that of the output vectors and change with the evenness of the frequency distribution: the more uniform is the distribution, the smaller should be the translation magnitude. Therefore, {the translation vector could be set as }
\bea\label{alpha}
T
= \alpha
\left(\begin{array}{c}
\hat n_1(F) \\
\hat n_2(F) \\
\cdots \\
\hat n_k(F)
\end{array} \right),~~
\alpha = \frac{\textrm{std}}{\textrm{std}_{\textrm{max}}} \left(o_{\textrm{max}}-o_{\textrm{min}}\right).
\eea
where
``$\textrm{std}_{\textrm{max}}$'' is the maximal standard deviation, which corresponds to the case where all samples fall into the same cluster, while ``$\textrm{std}$'' is the standard deviation of the current distribution.
$ \left(o_{\textrm{max}}-o_{\textrm{min}}\right)$ is the difference between the maximal and the minimal components in all outputs vectors, which gives the order of magnitude of $\alpha$. The ratio ``${\textrm{std}}/{\textrm{std}_{\textrm{max}}}$'' indicates the degree of unevenness: unevenness increases from 0 to 1 as ``$\textrm{std}$'' changes from 0 to the maximum. Noting that, from Equations (\ref{trans}) to (\ref{alpha}), all the quantities are from the outputs, no hyper--parameters are introduced. 

To make the distribution of pseudo--labels as uniform as possible, the translation should be repeatedly performed, and $\alpha$ should decrease as iterations of translation increase and, thus, we introduce a decay rate $\beta$ (see {Algorithm \ref{pcode})}. The pseudo--code of creating evenly distributed pseudo--labels by output translation is given in {Algorithm \ref{pcode}}.

\begin{algorithm}[H]
\label{pcode}
\caption{Creating even distributed pseudo--labels by output translation}
\KwIn{$N$ images $I(1), I(2),\cdots,I(N)$; cluster number $k$; decay rate $\beta$; $\alpha$ bound $\alpha_0$}
\KwOut{Even distributed pseudo--labels}
outputs = model(inputs)\;
labels = argmax(outputs)\;
C = count(labels) $-$ $N/k$\;
std = standard$\_$deviation(C)\;
\While{$\alpha>\alpha_0$  and  $\textrm{std}>0$}{
outputs = outputs - $\alpha$C\;
labels = argmax(outputs)\;
C = count(labels) $-$ $N/k$\;
std$\_$new = standard$\_$deviation(C)\;
\eIf{std$\_$new$<$std}{
std = std$\_$new\;
}{$\alpha = \alpha/\beta$\;
}
 }
\end{algorithm}

The translation (\ref{trans}) is just a simple linear transformation, that is, a smooth mapping 
\bq\label{o'o}
O'(s_1) \approx O'(s_2)~~\textrm{iff}~~O(s_1) \approx O(s_2).
\eq
Therefore, translation (\ref{trans}) does not change the smoothness of the current model, that is,~if
\bq\label{oi}
O(s_1) \approx O(s_2)~~\textrm{iff}~~I(s_1) \approx I(s_2),
\eq
then
\bq\label{o'i}
O'(s_1) \approx O'(s_2)~~\textrm{iff}~~I(s_1) \approx I(s_2).
\eq
More algorithm analysis can be found in Appendix {\ref{analysis}}, where we give a detailed comparison with the Sinkhorn--Knopp algorithm used in~\cite{conf/iclr/AsanoRV20a} in terms of of motivation and technical details, and where we also  give a clear mathematical illustration demonstrating how our algorithm preserves the distances of outputs. 
\vspace{6pt}

\vspace{6pt}

\textbf{Model update}. When the translation of outputs is completed, the $s$th input sample $I(s)$ is assigned to cluster $c_i$ according to (\ref{ci}), based on the final translated outputs $O'(s)=(o'_1(s),o'_2(s)\cdots,o'_k(s))$:
\bea\label{lbs}
I(s)\in c_i,\;\; y'(s) = (0, 0, \cdots, y'_i(s) = 1, \cdots, 0, 0)  \nonumber \\
\textrm{iff}~~o'_i(s) \geq o'_j(s) ~~\forall~~ j\neq i\;\;i,j = 1, 2,\cdots,k,~~~
\eea
where $y'(t)$ is the pseudo--label of $I(s)$. After reaching uniform labelling, we can update the weights of the network as standard supervised learning via standard cross--entropy loss
\bq
CE = - \frac{1}{N} \sum_s^N y'(s) \textrm{log} \;\big(\textrm{softmax}(O(s)\big),
\eq
which is established from the pseudo--labels $y'(t)$ and the outputs before translations $O(t)$.

The labels generated by (\ref{lbs}) from different versions of the same sample may be not consistent. For example, the grayscale image and the color image of the same sample will give different outputs, which may label the same image differently according to (\ref{lbs}). To remove this discrepancy, we propose a \emph{label--consistent--training (LCT)} method. In this method, we construct the loss function by the sum of the cross entropy losses as:
\bq\label{total-loss}
  L_{\textrm{LCT}}= \sum_{a,b}^{g}\left(- \frac{1}{N}\sum_{s}^{N} y'(s)_a \; \textrm{log}\big(\textrm{softmax}\left(O(s)_b\right)\big)\right).
\eq
In the loss above, $a,b$ indicates different transformations of the same images $I(s)$. If the labels from different versions of the same samples are not consistent, the loss function above would be hard to decrease. Thus, decreasing this loss enables different versions of the same samples to give the same prediction, i.e., the same pseudo--label, which will make the model smoother.  

\section{Experiments Resluts}\label{experiment}

In this section, we first give our experiment results on several datasets, which are often used in computer vision, to compare to the state--of--the--art research, and then we conduct experiments on various remote--sensing image datasets. 

{Recall that our method is an unsupervised representation learning method and its target is to learn good visual representations. Therefore, the ground--truth labels are never used in the training phase but the evaluation phase, and measuring the performance our method is to measure the performance of representations learned by our method. } To evaluate the learned representations, we consider a non--parametric and a parametric classifier: weighted kNN (k--nearest neighbors) and linear classifier~\cite{conf/cvpr/ZhangIE17}. {The evaluation process is supervised, that is, the labels are used, but only the CNN layers or the layers before the final layer are retained and frozen. }For the weighted kNN evaluation, we take  $k = 50$, $\sigma = 0.1$ and an embedding size of 128. For all experiments, we set decay rate to $\beta = 1.5$ and $\alpha_0=10^{-15}$ in  {Algorithm \ref{pcode}} .

\subsection{Compare to the State--of--the--Art Approaches}

To conveniently compare with other representation learning methods, in this part our experiments were conducted on a large--scale dataset, ImageNet LSVRC--12~\cite{imagenet_cvpr09} and three smaller scale datasets: CIFAR--10/100 and SVHN. ImageNet LSVRC--12 contains around 1.28 million training pictures of 1000 classes and 50--thousand validation pictures. There are 50,000 training images and 10,000 test images in the CIFAR--10/100 dataset, whose number of classes is 10/100 and pixel resolution is $32\times32$ . SVHN is similar to  {MNIST} (images of digits), which is a real--world image dataset. There are 73,257 images ($32\times32$) for training and 26,032 images ($32\times32$) for testing. All the ground truths of these datasets were used only in the evaluations of representations. The implementation details are given in Appendix  { \ref{im-details}}.

For fairly comparing to the previous work, we set the same cluster number $k$ for these datasets as \cite{conf/iclr/AsanoRV20a}. For CIFAR--10, the cluster number was set as 128. 
Therefore, the mean number of samples in each cluster is $50,000/128$. According to { Remark \ref{md}} , although this number is not an integer, setting it as the target still works: making the number of the samples assigned to each cluster as close as possible to the mean number. The cluster number is set as 128 for SVHN and 512 for CIFAR--100 when experiments were conducted in AlexNet.

Table \ref{ev-knn} gives kNN  and linear--classifier evaluations of representations learned from CIFAR--10/100 and SVHN through our method \textbf{OTL} (output translation). In addition, {kNN evaluations were performed based on the fully connected layer before the final layer, and all layers except the final layer were frozen. For linear classifier, all the fully connected layers were discarded and the retained CNN layers were frozen. The last CNN layer was then attached by a new randomly initialized fully connected layer whose weights were to be updated by a supervised learning.} With simple augmentations, the performances of our method reached the state--of--the--art performances. When strong augmentations and LCT (label consistent training) were employed, the performances of our method outperformed the previous works in the literature. For SVHN, the performance of our method is very close to the performance of supervised learning, especially linear classifier, where the accuracy is only $0.1\%$ lower than the accuracy in supervised learning. When one assumes the actual number of classes in the data is known, and sets $k$ equal to the actual number of classes, this can be regarded as giving the model some prior knowledge. It is seen that such prior knowledge can help improve the performances of our OTL algorithm, especially in the case for linear classiﬁer on SVHN, where the accuracy is the same as that of fully supervised learning. 

\begin{table}[H]
\caption{Evaluations on AlexNet ({OTL} = our new method). The best results are highlighted by the bold.}
\label{ev-knn}
\setlength{\tabcolsep}{5mm}
\newcolumntype{B}{>{\centering\arraybackslash}X}
\begin{tabularx}{\textwidth}{llll}
\toprule
       & \multicolumn{3}{l}{\textbf{kNN/FC}}  \\
         \midrule
  \textbf{Method} & \textbf{CIFAR--10} & \textbf{CIFAR--100} & \textbf{SVHN} \\
\midrule
  Supervised                                 & 91.9 & 69.7 & 96.5 \\
  Counting~\cite{conf/iccv/NorooziPF17}      & 41.7 & 15.9 & 43.4  \\
 DeepCluster~\cite{conf/eccv/CaronBJD18}     & 62.3 & 22.7 & 84.9 \\
  Instance~\cite{conf/cvpr/WuXYL18}          & 60.3 & 32.7 & 79.8 \\
  AND~\cite{conf/icml/HuangDGZ19}            & 74.8 & 41.5 & 90.9 \\
  SeLa~\cite{conf/iclr/AsanoRV20a}           & 77.6 & 44.2 & 92.8 \\ 
  ISL~\cite{wang2021instance}                & 82.8 & 50.3 & 91.0 \\
\midrule
\textbf{OTL}                               & 83.1 & 53.7 & 93.8  \\
\textbf{OTL}(Strong augmentations + LCT)             & \textbf{87.3} & \textbf{59.2} & \textbf{95.2}  \\
\midrule
\textbf{OTL}($k=$class number)             & 89.9 & 60.7 & 95.4  \\

\midrule
 & \multicolumn{3}{l}{Linear Classifier/conv5} \\
\midrule
  Supervised                                 & 91.8 & 71.0 & 96.1 \\
  Counting~\cite{conf/iccv/NorooziPF17}      & 50.9 & 18.2 & 63.4 \\
 DeepCluster~\cite{conf/eccv/CaronBJD18}     & 77.9 & 41.9 & 92.0 \\
  Instance~\cite{conf/cvpr/WuXYL18}          & 70.1 & 39.4 & 89.3 \\
  AND~\cite{conf/icml/HuangDGZ19}            & 77.6 & 47.9 & 93.7 \\
  SeLa~\cite{conf/iclr/AsanoRV20a}           & 83.4 & 57.4 & 94.5 \\ 
  ISL~\cite{wang2021instance}                & 85.8 & 60.1 & 93.9 \\
\midrule
\textbf{OTL}                               & 84.3 & 59.2 & 95.0 \\
\textbf{OTL} (Strong augmentations + LCT)            & \textbf{87.1} & \textbf{63.6} & \textbf{96.0} \\
\midrule
\textbf{OTL}($k=$class number)             & 90.8 & 65.6  & 96.1 \\
\bottomrule
\end{tabularx}
\end{table}

In Table \ref{ev-res}, the experiments were ran on ResNet--50 and for fairly comparing to other methods, $k=128$ was set for all experiments on ResNet--50. Both evaluations, kNN and linear classifier, were conducted on the last CNN layer. Note that both Instance~\cite{conf/cvpr/WuXYL18}, SimCRL~\cite{journals/corr/abs-2002-05709}  and ISL~\cite{wang2021instance} are methods of instance discrimination. This table shows that clustering--based representation learning methods could learn representations that are as good as that of instance discrimination or even better. Compared to Table \ref{ev-knn}, we find that the performances on ResNet--50 overall are better than on AlexNet, which is reasonable and predictable. 

\begin{table}[H]
\caption{Evaluations on ResNet--50 ({OTL} = Our new method). The best results are highlighted by the bold.}
\label{ev-res}
\begin{small}
\setlength{\tabcolsep}{10mm}
\newcolumntype{B}{>{\centering\arraybackslash}X}
\begin{tabularx}{\textwidth}{llll}
\toprule
       & \multicolumn{3}{l}{\textbf{kNN/FC}}  \\
         \midrule
  \textbf{Method} & \textbf{CIFAR--10} & \textbf{CIFAR--100} & \textbf{SVHN} \\
\midrule
  Instance~\cite{conf/cvpr/WuXYL18}          & 81.8 & 42.3 & 92.9 \\
  AND~\cite{conf/icml/HuangDGZ19}            & 87.6 & 49.0 & 93.2 \\
  ISL~\cite{wang2021instance}                & 88.9 & 58.1 & 94.5 \\
\midrule
\textbf{OTL}(ours)             & \textbf{90.0} & \textbf{66.4} & \textbf{95.2}  \\
\midrule
 & \multicolumn{3}{l}{Linear Classifier/conv5} \\
\midrule
  Instance~\cite{conf/cvpr/WuXYL18}          & 85.0 & 50.1 & 94.4 \\
  AND~\cite{conf/icml/HuangDGZ19}            & 90.2 & 58.2 & 94.9 \\
  SimCRL~\cite{journals/corr/abs-2002-05709} & 90.6 & \textbf{71.6} & -  \\
  ISL~\cite{wang2021instance}                & 91.5 & 65.9 & 95.2 \\
\midrule
\textbf{OTL} (ours)             & \textbf{92.0} & 70.7 & \textbf{95.5} \\
\bottomrule
\end{tabularx}
\end{small}
\end{table}
To demonstrate the effectiveness of our method on a large--scale dataset and the speed of our algorithm in dealing with huge number of images, i.e., over one million, we also conducted experiments on ImageNet with AlexNet. As can be seen from {Table} \ref{img-eval}, our method, without strong augmentations and LCT, achieves state--of--the--art performances in evluations of both linear and kNN classifiers. In our model, comparing the representations from the five CNN layers, representations from the last two layers perform better than that of the first three layers. 
\begin{table}[h]
\caption{Evaluation on ImageNet. ``$^*$'' denotes training on larger AlexNet.''3k $\times$ 10/1'' denotes 3000 clusters and 10/1 heads (10/1 fully connected layers attached at the end of the architecture). The best results are highlighted by the bold.\label{img-eval}}
\begin{small}
\setlength{\tabcolsep}{3mm}
\newcolumntype{B}{>{\centering\arraybackslash}X}
\begin{tabularx}{\textwidth}{lllllll}
\toprule
  \textbf{Classifier} & \multicolumn{5}{c}{\textbf{Linear Classifier}} & kNN \\
 \midrule
  \textbf{Feature} & \textbf{conv1} & \textbf{conv2} & \textbf{conv3} & \textbf{conv4} & \textbf{conv5} & \textbf{FC} \\
\midrule
 Supervised \cite{conf/cvpr/ZhangIE17} & 19.3 & 36.3 & 44.2 & 48.3 & 50.5 & -\\
  Random \cite{conf/cvpr/ZhangIE17} & 11.6 & 17.1 & 16.9 & 16.3 & 14.1 & 3.5\\
  Inpainting \cite{conf/cvpr/PathakKDDE16} & 14.1 & 20.7 & 21.0 & 19.8 & 15.5 & - \\
  BiGAN \cite{conf/iclr/DonahueKD17} & 17.7 & 24.5 & 31.0 & 29.9 & 28.0 & -\\
  Instance retrieval \cite{conf/nips/XieDDHN17} & 16.8 & 26.5 & 31.8 & 34.1 & 35.6 & 31.3 \\
  RotNet \cite{GoodBengCour16} & 18.8 & 31.7 & 38.7 & 38.2 & 36.5 & -\\
  AND $^*$ \cite{conf/cvpr/JenniF18} & 15.6 & 27.0 & 35.9 & 39.7 & 37.9 & 31.3\\
  CMC $^*$ \cite{JMLR:v11:vincent10a} & 18.4 & 33.5 & 38.1 & 40.4 & 42.6 & -\\
  AET $^*$ \cite{conf/cvpr/YeZYC19} & 19.3 & \textbf{35.4} & \textbf{44.0} & 43.6 & 42.4 & -\\
  SeLa[3k$\times$10] $^*$ \cite{conf/iclr/AsanoRV20a} & 20.3 & 32.2 & 38.6 & 41.4 & 39.6 & -\\
 ISL~\cite{wang2021instance} & 17.3 & 29.0 & 38.4 & 43.3 & 43.5 & \textbf{38.9} \\
 \midrule
  \textbf{OTL[3k$\times$1] $^*$} (ours) & \textbf{20.6} & 34.0 & 41.2 & \textbf{44.6} & \textbf{43.8} & 36.7 \\
\bottomrule
\end{tabularx}
 \end{small}
\end{table}

\subsection{Impact of Even Pseudo--Label Distribution on Performances}


We have to emphasize, again, that evenly distributing pseudo--labels is in order to make the representations highly discriminative. Besides even distribution, our algorithm can conveniently set various unevenly distributed pseudo--labels, which can be performed by modifying the $\bar n(F)$ in frequency indicator $C$ (\ref{indicator}) as the desired distribution.

\textbf{Unevenly distributed pseudo--label}. To investigate the influences of unevenness of pseudo--labels' distribution, various uneven target distributions of pseudo--labels were set, as shown in Figure \ref{unp} (for more details, see Appendix  {\ref{un-plbs})} for training on CIFAR--10/100. The kNN evaluations for different target distributions are given in Table \ref{dis-com} (these models were trained on AlexNet with fewer epochs than in Table \ref{ev-knn} and without strong augmentations and the LCT technique). As one can see from both tables, the accuracy of kNN decreases as evenness decreases. This is consistent with the argument made in the method section that evenly distributed pseudo--labels are good for learning representations. 
\vspace{-9pt}
\begin{figure}[h]
\centering
\subfigure {
\includegraphics[width=3.0in]{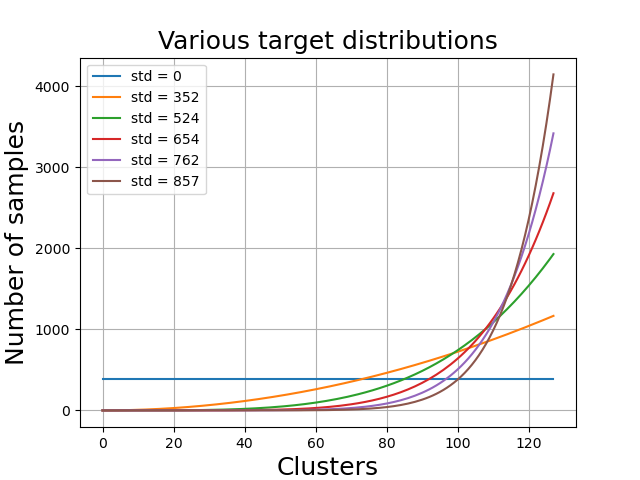}
}
\subfigure {
\includegraphics[width=3.0in]{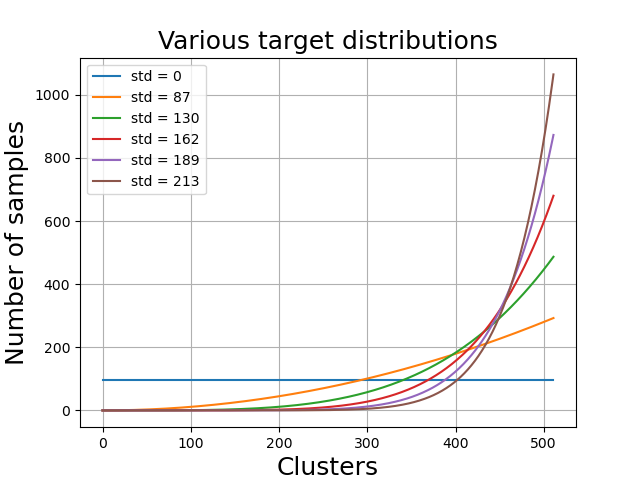}
}
\caption{Left: various target distributions of pseudo--labels for dataset CIFAR--10; Right:
various target distributions of pseudo--labels for dataset CIFAR--100.}
\label{unp}
\end{figure}\vspace{-12pt}
\begin{table}[H]
\caption{CIFAR--10/100: evaluations for unevenly distributed pseudo--labels. The best results are highlighted by the bold.}
\label{dis-com}

\setlength{\tabcolsep}{6mm}
\newcolumntype{B}{>{\centering\arraybackslash}X}
\begin{tabularx}{\textwidth}{lllllll}
\toprule
\multicolumn{7}{l}{CIFAR--10}\\
\midrule
std & 0 & 352.3 & 524.4 & 654.4 & 654.3 & 857.5 \\
kNN & \textbf{79.4} & 78.2 & 77.7 & 77.3 & 76.1 & 75.4 \\
\midrule
\multicolumn{7}{l}{CIFAR--100}\\
\midrule
 std & 0 & 87.5 & 130.4 & 162.8 & 189.8 & 213.4 \\
 
 kNN & \textbf{48.9} & 48.0 & 46.5 & 44.72 & 45.4 & 44.9 \\
\bottomrule
\end{tabularx}
\end{table}

\textbf{Unevenly distributed datesets}. The standard datasets used in this paper are evenly distributed, i.e., each class contains the same number of images. However, the effectiveness of our method does not depend on the data being evenly distributed. In order to show that, we considered five unevenly distributed datasets which are made from CIFAR--10 by deleting different numbers of images from different classes. As a comparison, five evenly distributed datasets were made, and each has the same number of samples as  the corresponding dataset in the five unevenly distributed ones.

In Table \ref{imb-data} (again with fewer epochs and without strong augmentations and the LCT technique), for unevenly distributed ``D$\_$100'', we deleted 0, 100, 200, $\cdots$ 900 images  from class 0, 1, 2, $\cdots$, 9, while evenly distributed ``D$\_$100'' has the same total number of images but the ground truth is evenly distributed. Similarly, for unevenly distributed ``D$\_$200'', we deleted 0, 200, 400, $\cdots$ 1800 images from class 0 to 9 . For unevenly distributed data ``D$\_$500'', the number of images in the first class is ten times the number of images in the last class.

We can see from Table \ref{imb-data} that our method is almost unaffected by whether the actual data distribution is even or not. As can be expected, evenly distributed data leads to slightly better performances than that on the unevenly distributed data. This experiment demonstrates that the success of our algorithm does not rely on the even distribution of the~datasets.

\begin{table}[h]
\caption{Unevenly VS evenly distributed samples: kNN evaluation.}
\label{imb-data}
\setlength{\tabcolsep}{6mm}
\newcolumntype{B}{>{\centering\arraybackslash}X}
\begin{tabularx}{\textwidth}{llllll}
\toprule
\textbf{Dataset} & \textbf{D}$\_$\textbf{100} & \textbf{D}$\_$\textbf{200} & \textbf{D}$\_$\textbf{300} & \textbf{D}$\_$\textbf{400} & \textbf{D}$\_$\textbf{500}  \\
\midrule
   Uneven & 77.0 & 76.8 & 76.0 & 75.4 & 73.6 \\
   Even   & 77.4 & 77.0 & 76.8 & 75.8 & 74.5 \\
\bottomrule
\end{tabularx}
\end{table}

\subsection{Application to remote--sensing Images}

For all remote--sensing datasets, we divided images into two groups: the training sets and testing sets, that is, we randomly picked around 80$\%$ images for training and the others for testing. All training only involved training sets without using any labels. For all datasets, we employed the same backbone, i.e., ResNet-18. All the remote--sensing images were resized to 256 $\times$ 256 and then cropped to 224 $\times$ 224 and for the strong augmentations we used the strategy from \cite{conf/eccv/GansbekeVGPG20} with Augment = 10, n$_{holes}$ = 1, and length = 16. In this part, we will apply our method to seven remote--sensing image datasets: HMHR (How to Make High Resolution Remote Sensing Image dataset), EuroSAT, SIRI WHU google, UCMerced LandUse, AID, PatternNet, NWPU RESISC45.

For the very small datasets---HMHR, SIRI WHU google, UCMerced LandUse---we used pretrained models. The training epoch for these three datasets is 64, and we set initial learning rate as 0.03 and drops to 0.003, 0.0006 at epoch 32, 48. For the training without a pretrained model, the total training epoch was set to 300, and learning rate initially was 0.03 which dropped to 0.003 and 0.0006 at epoch 160 and 240. For all the training, the momentum was set to 0.9. For all datasets, the $k$ was set to 128.

This is a dataset that is made from google map via LocalSpece Viewer. In this dataset, there only contains 533 images with five classes: building, farmland, greenbelt, wasteland, water. The resolution and spatial resolution of the remote--sensing images are 256 $\times$ 256 and 2.39 m. To prevent the network from overfitting, we used the strong augmentations \cite{conf/eccv/GansbekeVGPG20} and the pretrained models trained from other remote--sensing images datasets: PatternNet, NWPU RESISC45 and EuroSAT. All the pretained models were trained through our method discussed in the Methods section and no labels were used in the training. For supervised training, we only considered the pretrained model on PatternNet. From Table \ref{hrs}, 
we see that our method works well on very small datasets with few classes. There is only a small gap ($2.8\%$) between supervised method and ours in linear classifier evaluation. Although using different pretrained models has different performances, all of them  are effective and perform well. 
\begin{table}[h]
\caption{HMHR: How-to-make-high-resolution-remote-sensing-image-dataset.}
\label{hrs}
\setlength{\tabcolsep}{2mm}
\newcolumntype{B}{>{\centering\arraybackslash}X}
\begin{tabularx}{\textwidth}{llllllll}
\toprule
\textbf{Method} & \textbf{k} & \textbf{supervised} & \textbf{OTL (ours)}  & & \\
\textbf{Pretrained on} &  & \textbf{PatternNet} & \textbf{PatternNet} & \textbf{NWPU RESISC45} & \textbf{EuroSAT} \\
\midrule
    kNN & 10  & 82.9 &  79.0 & 75.2 & 72.4 \\
    & 50 & 83.8 & 78.1 & 76.2  & 73.3 \\
    & 100 & 84.8 & 78.1 & 78.1 & 76.2 \\
    & 200 & 83.8 & 77.1  & 79.0 & 75.2\\
   \midrule
   Linear Classifier   & - &  88.5  & 85.7 & 83.8 & 81.9 \\
\bottomrule
\end{tabularx}
\end{table}

EuroSAT \cite{helber2019eurosat,helber2018introducing} is a dataset that for land use and land cover classification. It is consist of 10 classes with 27,000 labeled images of annual crop, forest, highway, reiver, sealake and so on. The resolution and spatial resolution of these images are 64 $\times$ 64 and 10 m. Table \ref{eur} shows that the gap between supervised method and ours is only $1\%$ in Linear classifier evaluation. In kNN evaluation, the performance of our method is also very close to the performance of supervised learning. 

\vspace{-9pt}
\begin{table}[H]
\caption{EuroSAT : Land Use and Land Cover Classification with Sentinel-2. 
}
\label{eur}
\setlength{\tabcolsep}{9mm}
\newcolumntype{B}{>{\centering\arraybackslash}X}
\begin{tabularx}{\textwidth}{llllll}
\toprule

\textbf{Method} & \textbf{k} & \textbf{supervised} & \textbf{OTL (ours)}  \\
\midrule
    kNN & 10  & 96.8 & 94.4 \\
    & 50 & 96.6 & 94.7 \\
    & 100 & 96.3 & 94.6  \\
    & 200 & 95.9 &  94.4 \\
   \midrule
   Linear Classifier   & - &  96.9  &  95.9\\
\bottomrule
\end{tabularx}
\end{table}

The dataset of SIRI WHU google \cite{ma2015adaptive} contains 12 classes images with 2400 images. There are 200 images for each class. The class of images includes: agriculture, commercial, harbor, idle land, industrial, meadow, overpass, park, pond, river, water, residential. The pixel and spatial resolution of images in this dataset are $200\times200$ and 2 m. As can be seen from Table \ref{whu}, using the pretained model on NWPU RESISC45 have the best performance which even is better than the that of supervised learning based on pretrained model on PatternNet. Representations learned based on pretrained model on PatternNet perform closely to that of supervised learning. 

\begin{table}[H]
\caption{SIRI WHU google.}
\label{whu}
\setlength{\tabcolsep}{2mm}
\newcolumntype{B}{>{\centering\arraybackslash}X}
\begin{tabularx}{\textwidth}{llllllll}
\toprule
\textbf{Method} & \textbf{k} & \textbf{supervised} & \textbf{OTL (ours)}  &  &  \\
\textbf{Pretrained on} &  & \textbf{PatternNet} & \textbf{PatternNet} & \textbf{NWPU RESISC45} & \textbf{EuroSAT} \\
\midrule
    kNN & 10  & 86.9 & 85.4  & 89.0 & 80.8\\
    & 50 & 83.8 & 82.5  & 85.6 & 77.9\\
    & 100 & 83.5 & 80.4 & 84.6 & 76.0 \\
    & 200 & 81.3 & 79.0  & 83.3 & 73.8\\
   \midrule
   Linear Classifier   & - &  89.6  & 88.6  & 89.7 & 85.7\\
\bottomrule
\end{tabularx}

\end{table}

UCMerced landUse \cite{yang2010bag} gives the images that were manually extracted from large images from the USGS National Map Urban Area Imagery collection for various urban areas around the country. Each image measures 256 $\times$ 256 pixels, with a 0.3 m spatial resolution. There are 21 classes with 2100 images, including airplane, beach, buildings, freeway, golfcourse, tenniscourt and so on and there are 100 images for each class.  Table \ref{uc} demonstrates that the performance of our method based on pretrained models on PatternNet and NWPU RESISC45 is very close to supervised learning, and in Linear Classification, our method even perform better. 

\begin{table}[H]
\caption{UCMerced LandUse. 
}
\label{uc}
\setlength{\tabcolsep}{1.5mm}
\newcolumntype{B}{>{\centering\arraybackslash}X}
\begin{tabularx}{\textwidth}{lllllll}
\toprule
\textbf{Method} & \textbf{k} & \textbf{supervised} & \textbf{OTL (ours)} &  \\
\textbf{Pretrained on} &  & \textbf{PatternNet} & \textbf{PatternNet} & \textbf{NWPU RESISC45} & \textbf{EuroSAT} \\
\midrule
    kNN & 10  & 88.1 & 86.9 & 86.2 & 81.7\\
    & 50 & 87.4 & 87.9 & 86.4 & 82.4\\
    & 100 & 87.1 & 88.1 & 86.2 & 82.6\\
    & 200 & 86.2 & 88.6 & 86.2 & 82.6\\
   \midrule
   Linear Classifier   & - &  89.6  &  90.2 & 89.7 & 87.2\\
\bottomrule
\end{tabularx}
\end{table}

AID \cite{xia2017aid} is a aerial image dataset with high resolution, 600 $\times$ 600 pixels, and a \mbox{0.5--0.8 m} spatial resolution, which constitutes collected sample images from Google Earth imagery. There are 30 classes with 10,000 images. This dataset is not evenly distributed and in each class there are about 200 to 400 images. The scene classes include bare land, baseball field, beach, bridge, center, church, dense residential, desert,  forest, meadow, medium residential, mountain, park, parking, playground, port, railway station, resort, school and so on. This experiment shows that, for high-resolution remote--sensing images and an unevenly distributed remote--sensing dataset, our method also works well (see table \ref{aid}). 

\begin{table}[H]
\caption{AID: a benchmark dataset for performance evaluation of aerial scene classification. 
}
\label{aid}
\setlength{\tabcolsep}{9mm}
\newcolumntype{B}{>{\centering\arraybackslash}X}
\begin{tabularx}{\textwidth}{llllll}
\toprule
\textbf{Method} & \textbf{k} & \textbf{supervised} & \textbf{OTL (ours)}   \\
\midrule
    kNN & 10  & 87.8 &  83.6\\
    & 50 & 87.6 & 83.5 \\
    & 100 & 86.9 & 82.6 \\
    & 200 & 86.7 & 81.9 \\
   \midrule
   Linear Classifier   & - &  89.5  &  85.5\\
\bottomrule
\end{tabularx}
\end{table}

PatternNet \cite{zhou2018patternnet} is a remote--sensing dataset with 30,400 high-resolution images whose resolution is 256 $\times$ 256 pixels and spatial resolution is 0.062--4.693 m. These images were collected for remote--sensing image retrieval from Google Earth imagery or via the Google Map API for some US cities. For each of the 38 classes, there are 800 images. From the results in Table \ref{pat}, we see that, in kNN evaluation, our method performs as well as that of supervised learning; while in linear classifier evaluation, our method has an even better performance. 

\begin{table}[H]
\caption{PatternNet.}
\label{pat}
\setlength{\tabcolsep}{9mm}
\newcolumntype{B}{>{\centering\arraybackslash}X}
\begin{tabularx}{\textwidth}{llllll}
\toprule
\textbf{Method} & \textbf{k} & \textbf{supervised} & \textbf{OTL (ours)}  \\
\midrule
    kNN & 10  & 96.5 & 96.4 \\
    & 50 & 96.2 & 96.1 \\
    & 100 & 95.7 & 95.7  \\
    & 200 & 95.5 &  95.6 \\
   \midrule
   Linear Classifier   & - &  96.7  &  97.4 \\
\bottomrule
\end{tabularx}
\end{table}

NWPU RESISC45 \cite{cheng2017remote} is made by Northwestern Polytechnical University (NWPU), which is avaiable for remote--sensing image scene classification (RESISC). It contains 31,500 images in total with a pixel resolution of 256 $\times$ 256 and a 0.2--30 m spatial resolution. This dataset cover 45 scene classes with 700 images in each class. These 45 scene classes include  baseball diamond, basketball court,  bridge, chaparral, church, circular-farmland, cloud, commercial area, dense residential, desert, intersection, island, lake, meadow, medium residential, mobile home park, wetland and so on. The experiment in Table \ref{npu} shows that, for a dataset with more classes, our method still has an excellent performance that is very close to performance of supervised learning. 

\begin{table}[H]
\caption{NWPU RESISC45.}
\label{npu}
\setlength{\tabcolsep}{9mm}
\newcolumntype{B}{>{\centering\arraybackslash}X}
\begin{tabularx}{\textwidth}{llllll}
\toprule
\textbf{Method} & \textbf{k} & \textbf{supervised} & \textbf{OTL (ours)}  \\
\midrule
    kNN & 10  & 92.1 & 89.3 \\
    & 50 & 90.6 & 88.7 \\
    & 100 & 89.7 & 88.3 \\
    & 200 & 89.1 & 87.4 \\
   \midrule
   Linear Classifier   & - &  92.3  & 89.8  \\
\bottomrule
\end{tabularx}
\end{table}

From Tables \ref{hrs} to \ref{npu}, we see that, for the pixel resolutions of remote--sensing images ranging from $64\times64$ to $600\times600$, i.e., from low to high resolution, spatial resolutions ranging from 0.062 m to 30 m, class numbers ranging from 5 to 45, and total numbers of images ranging from 533 to 31,500, our method can learn good representations, which are close to, or even better than, the representations learned from supervised learning. These experiments demonstrate that our method could be widely applied to learn the representations of remote--sensing images. 

\begin{table}[H]
\caption{Five different ways to divide the dataset SIRI WHU google.}
\label{5whu}
\setlength{\tabcolsep}{4mm}
\newcolumntype{B}{>{\centering\arraybackslash}X}
\begin{tabularx}{\textwidth}{lllllll}
\toprule
\textbf{Dataset} & \textbf{Evaluation method} & \textbf{k} & \textbf{supervised} & \textbf{OTL (ours)}  \\
\midrule
   Division 1 & kNN      & 10  & 86.9 & 85.4 \\
    &                    & 50  & 83.8 & 82.5 \\
    &                    & 100 & 83.5 & 80.4 \\
    &                    & 200 & 81.3 & 79.0 \\
  &  Linear Classifier   & N/A & 89.6 & 88.6 \\
  \midrule
   Division 2 & kNN      & 10  & 92.5 & 89.6 \\
                  &      & 50  & 91.0 & 89.4 \\
                  &      & 100 & 89.4 & 87.9 \\
                  &      & 200 & 88.3 & 87.3 \\
  &  Linear Classifier   & N/A & 92.6 & 92.1 \\
    \midrule
   Division 3 & kNN      & 10  & 93.2 & 88.8 \\
                  &      & 50  & 90.6 & 88.3 \\
                  &      & 100 & 88.8 & 87.3 \\
                  &      & 200 & 88.5 & 87.1 \\
  &  Linear Classifier   & N/A & 93.0 & 92.7 \\
      \midrule
   Division 4 & kNN      & 10  & 93.1 & 89.8 \\
                  &      & 50  & 90.6 & 89.4 \\
                  &      & 100 & 89.4 & 87.9 \\
                  &      & 200 & 88.5 & 86.3 \\
  &  Linear Classifier   & N/A & 92.8 & 92.5 \\
        \midrule
   Division 5 & kNN      & 10  & 92.3 & 90.4 \\
                  &      & 50  & 90.4 & 89.0 \\
                  &      & 100 & 89.0 & 88.1 \\
                  &      & 200 & 87.9 & 87.1 \\
  &  Linear Classifier   & N/A & 92.8 & 92.3 \\
\bottomrule
\end{tabularx}
\end{table}

To investigate the effect of choosing different training and testing samples in the experiments of remote--sensing images, we conduct  four more contrast experiments on the dataset SIRI WHU google. In this dataset, there are only 2400 images, and, thus, the random choice of training and testing samples may be quite different. We randomly selected $80\%$ images as training samples and $20\%$ as testing samples four times to construct four different divisions of the dataset SIRI WHU google. The performances on the five different divisions are given in Table \ref{5whu}. From Table \ref{5whu}, we see that different training and testing samples would affect the performances both on supervised learning and our method. However, in all five cases, using our method, without any labels can always learn good representations for remote--sensing images.

\vspace{-9pt}
\begin{figure}[H]
\setlength{\abovecaptionskip}{0.cm}
\setlength{\belowcaptionskip}{-0.cm}
\centering
\subfigure[ ] {
 \label{fig:a}
\includegraphics[width=3.0in]{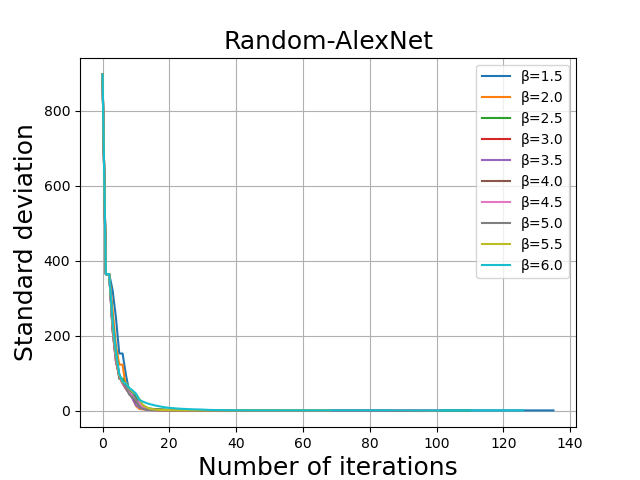}
}
\subfigure[ ] {
\label{fig:b}
\includegraphics[width=3.0in]{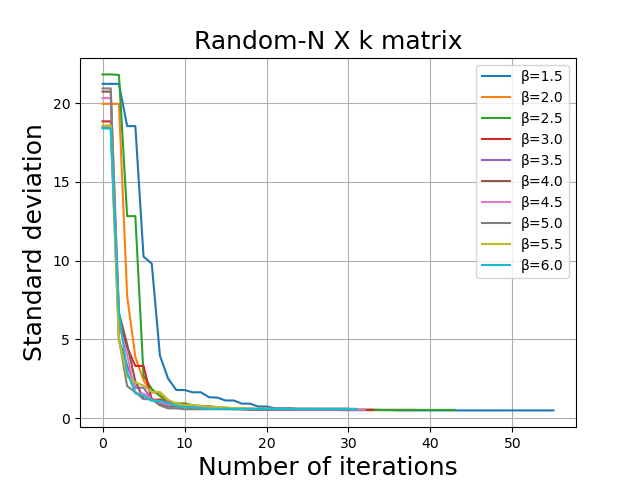}
}
\subfigure[ ] {
\label{fig:c}
\includegraphics[width=3.0in]{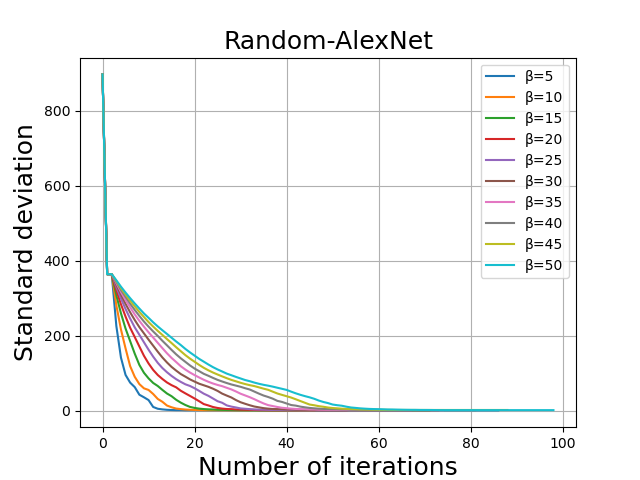}
}
\subfigure[ ] {
\label{fig:d}
\includegraphics[width=3.0in]{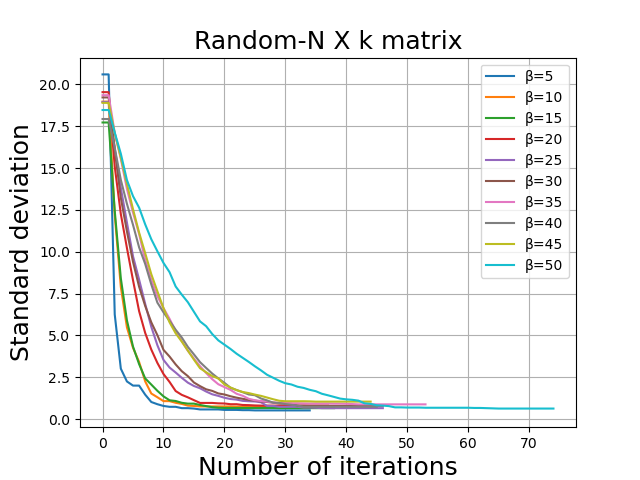}
}
\caption{For $k=128$, the performance of our method with various decay rate $\beta$ in the Algorithm  \ref{pcode} is given. The x axis is the number of iterations and y axis is the standard deviations for the distribution of pseudo--labels. In (\textbf{a},\textbf{c}) the method is applied on the outputs of CIFAR--10 from randomly weighted AlexNet; in (\textbf{b},\textbf{d}) the method is applied on $N\times k$ matrices randomly created.}
\label{alpha_0}
\end{figure}

\section{Discussion}
To study the robustness of our algorithm, we consider how the two parameters, cluster number $k$ and decay rate $\beta$ (see  {Algorithm \ref{pcode}}), affect the performance of our algorithm. For given $N$, $k$ and $\alpha_0$, the decay rate is the unique parameter in our algorithm, and has to be larger than one such that the magnitude of translations will decrease as iterations increase. We conducted two groups of experiments for $\beta$ ranging from 1.5 to 6 and from 5 to 50 in Figure \ref{alpha_0}, and the results show that our algorithm can effectively create evenly distributed pseudo--labels for these wide ranges of $\beta$ values, indicating our algorithm is insensitive to this free parameter. 

\begin{figure*} [h]
\centering
\subfigure[ ] {
 \label{fig:a}
\includegraphics[width=3.0in]{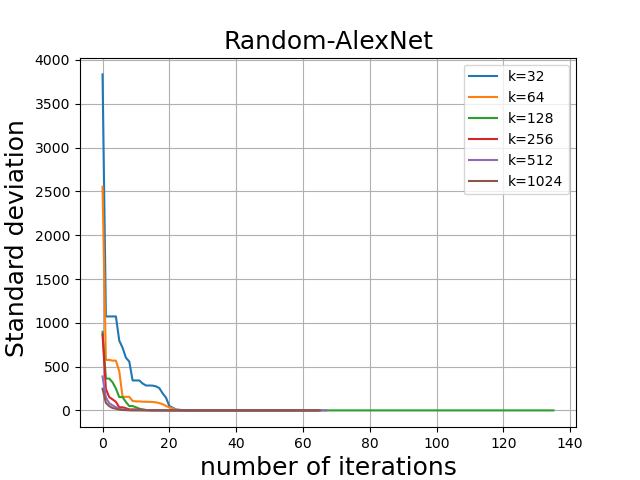}
}
\subfigure[ ] {
\label{fig:b}
\includegraphics[width=3.0in]{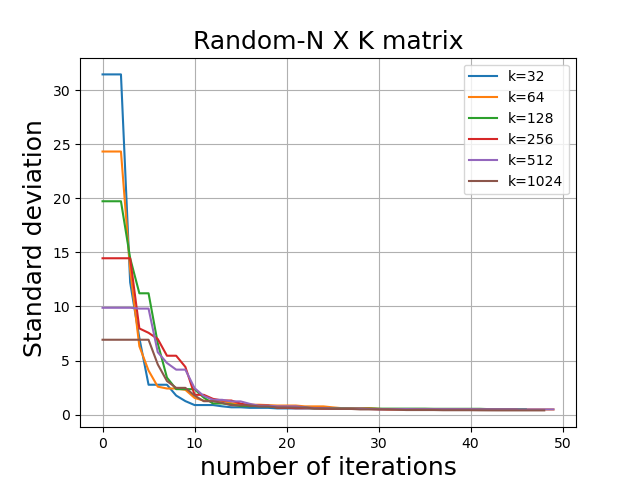}
}
\caption{For $N$ = 50,000 and $\beta = 1.5$, the performance of our algorithm under various cluster number $k$. In both (\textbf{a},\textbf{b}), for all $k$, convergence needs only around 20 iterations or fewer and the final distributions are quite uniform.}
\label{nk}
\vspace{-0.8em}
\end{figure*}

In Figure \ref{nk}, we demonstrate that our method is not only effective but also efficient (it only needs around 20 iterations to converge) in various settings of cluster number $k$. In Tables \ref{diff-k-10} and \ref{diff-k-100} (without LCT and strong augmentations), we show the kNN evaluations for the representations learned using different $k$, which demonstrates that our method works well for $k$ ranging from 32 to 2048. When $k=N$, our method would be the same as instance discrimination. Increasing $k$ would make representations more discriminative, while decreasing $k$ would make representations closer. Representation learning requires that representations of different samples to be as distinct as possible while representations of similar samples get closer. Therefore, very large or small $k$ are not considered in this paper: small $k$ would not give the highly discriminative representations while large $k$ would only consider the nearest neighbors as the same class, or even only the transformations of the same sample as instance discrimination does. 

\begin{table}[H]
\caption{CIFAR--10: Evaluations for different $k$.}
\label{diff-k-10}
 \begin{sc}
\setlength{\tabcolsep}{6mm}
\newcolumntype{B}{>{\centering\arraybackslash}X}
\begin{tabularx}{\textwidth}{lllllll}
\toprule
   $k$& \textbf{32} & \textbf{64} & \textbf{128} & \textbf{256} & \textbf{512} & \textbf{1024} \\
   \midrule
   $\bar n$ & 1562.5 & 718.3 & 390.6 & 195.3 & 97.7 & 48.8 \\
   kNN & 77.1 & 78.6 & 79.4 & 79.0 & 78.9 & 77.8 \\
\bottomrule
\end{tabularx}
 \end{sc}
\end{table}\vspace{-9pt}
\begin{table}[H]
\caption{CIFAR--100: Evaluations for different $k$.}
\label{diff-k-100}
 \begin{sc}
\setlength{\tabcolsep}{6mm}
\newcolumntype{B}{>{\centering\arraybackslash}X}
\begin{tabularx}{\textwidth}{lllllll}
\toprule
$k$ & \textbf{64} & \textbf{128} & \textbf{256} & \textbf{512} & \textbf{1024} & \textbf{2048} \\
\midrule
  $\bar n$ & 718.3 & 390.6 & 195.3 & 97.7 & 48.8 & 24.4 \\
   kNN & 43.8 & 46.1 & 47.8 & 48.9 & 49.5 & 48.9 \\
\bottomrule
\end{tabularx}
 \end{sc}
\end{table}

\begin{figure}[H]
\centering
\subfigure[ ] {
 \label{fig:a}
\includegraphics[width=3.0in]{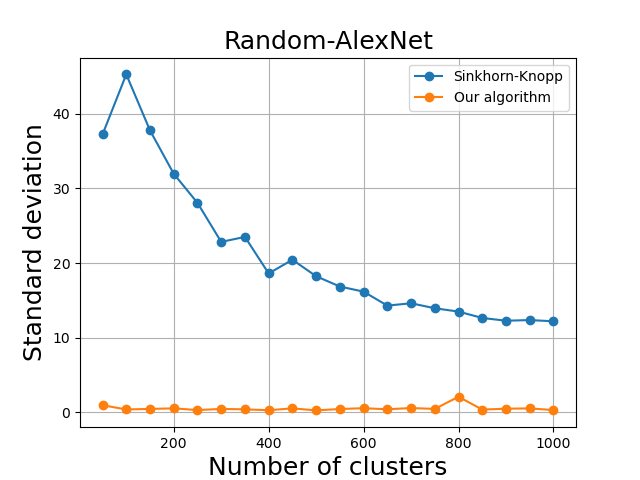}
}
\subfigure[ ] {
\label{fig:b}
\includegraphics[width=3.0in]{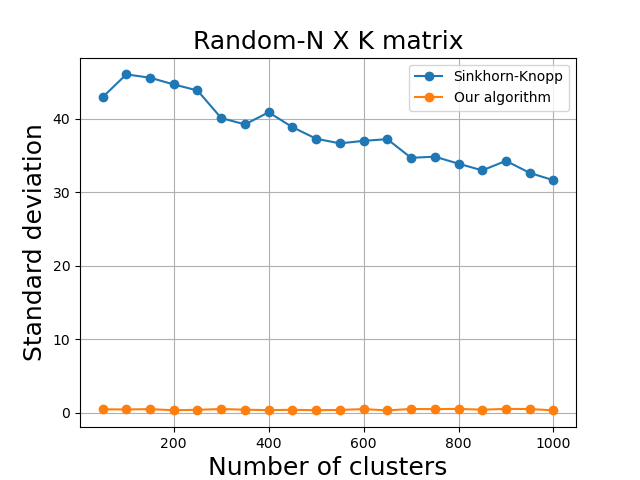}
}
\caption{For $N$ = 50,000, comparisons between the Sinkhorn--Knopp algorithm and our algorithm: we consider 20 different $k$ ($50 , 100, 150,\cdots, 1000$) for two kinds of $N\times K$ matrices: one is output of the randomly initialized AlexNet with CIFAR--10; one is randomly created $N\times k$ matrix. In both (\textbf{a},\textbf{b}), for all $k$, our algorithm has better performance on approaching even distributions.}
\label{skvsfs}
\end{figure}

The comparisons between our method and the Sinkhorn--Knopp algorithm are given in Figure \ref{skvsfs}. The Sinkhorn--Knopp algorithm is a classical method which has been widely used, especially in transport problems. For 20 different $k$ values and different $N\times K$ matrices, our method always obtained more uniform pseudo--labels (see Figure \ref{skvsfs}).\vspace{-9pt}

\begin{figure}[H]
\centering
\subfigure[ ] {
 \label{fig:a}
\includegraphics[width=3.0in]{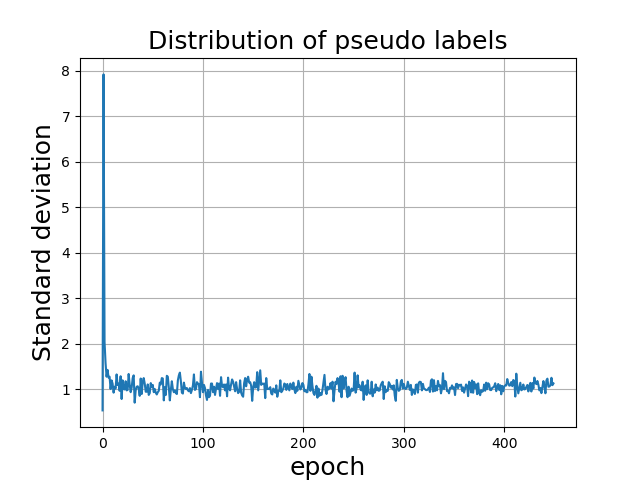}
}
\subfigure[ ] {
\label{fig:b}
\includegraphics[width=3.0in]{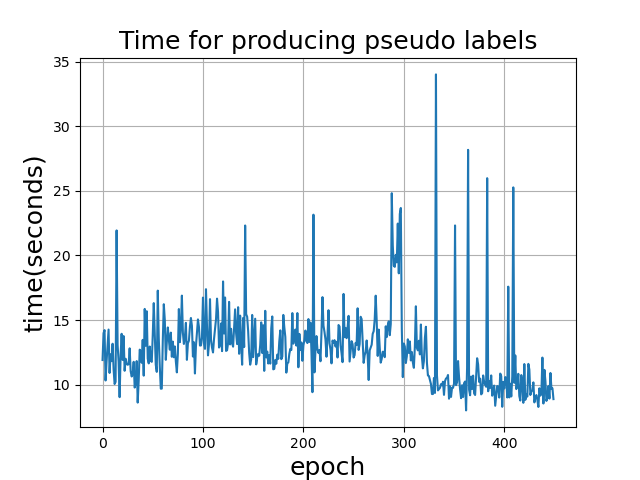}
}
\caption{For $k=3000$ and 450 epochs on AlexNet, the standard deviation of pseudo--labels and time for convergence at every epoch: (a)  the very small values of standard deviation means that the distribution of pseudo--labels is very uniform; (b) the average time to produce these labels is 12.68 s on GPU.}
\label{img}
\end{figure}

The main advantage of the Sinkhorn--Knopp algorithm is its speed of convergence. For creating pseudo--labels from outputs of ImageNet, the Sinkhorn--Knopp algorithm converges within 2 min~\cite{conf/iclr/AsanoRV20a}. In Figure \ref{img}, we give the standard deviations of pseudo--labels for ImageNet and the time for producing them at each epoch by our method. As can been seen from Figure \ref{img}a, for all epochs, the pseudo--labels created from our method are very uniform and the average standard deviation is only 1.07. From (b) in Figure \ref{img}, we see that our algorithm is very fast to converge even for 1.28 million pictures and 3000 clusters. The average time for approaching even distribution is only 12.68 s on GPU (NVIDIA A100).\vspace{-9pt}

\section{Conclusions}\label{conclusions}

In this paper, under a clustering model where images are clustered based on the representations of the images, we found a quantity that can be used to compare the discriminativeness of the representations and, based on this quantity, we demonstrated that evenly distributing images into clusters requires their representations to be most discriminative. We developed an algorithm to translate outputs such that pseudo--labels can be created evenly according to representations of images while keeping the smoothness of the model unchanged. Extensive experimental results  demonstrate that our method is very effective and can learn representations that are as good as or better than the state--of--the--art methods. We then applied our method to various remote--sensing image datasets and the representations learned by our method are very close to, or even better than, the representations learned by supervised learning.

\vspace{6pt}

\appendix
\section{Examples for Measuring the Discriminativeness of Representations}\label{sim-exam}

\subsection{Four Images Clustering}

In this section, we give a simple toy model: four images clustered through the model, as in Figure \ref{cl_rp}. If we set the cluster number to four, all five possible ways to cluster are given in Table \ref{tbcl4}. Case 1 in Table \ref{tbcl4} (there are four images in cluster C1 while no images are assigned to other clusters) corresponds to the degenerated solution, which is given by the most indiscriminative representations. This case gives the most indistinguishable pairwise comparisons, i.e., 6. The number of indistinguishable pairwise comparisons decreases from case 1 to case 5. For the last case, there are the most distinguishable pairwise comparisons, which requires the representations to be the most discriminative. This is reasonable when we remember that all the images are different: the most discriminative representations should reveal the most differences in the images. In fact, the last case is exactly what instance discrimination does. 
\begin{table}[H]
\caption{Five possible clusterings with cluster number as four. ``IND'': the number of indistinguishable pairwise comparisons; “DIS'': the number of distinguishable pairwise comparisons. }
\label{tbcl4}
\setlength{\tabcolsep}{6mm}
\newcolumntype{B}{>{\centering\arraybackslash}X}
\begin{tabularx}{\textwidth}{lllllll}
\toprule
 \textbf{Cluster} & \textbf{C1} & \textbf{C2} & \textbf{C3} & \textbf{C4} & \textbf{Ind} & \textbf{Dis} \\
\midrule
  Case 1  & 4 & 0 & 0 & 0 & 6 & 0 \\
  Case 2  & 3 & 1 & 0 & 0 & 3 & 3 \\
  Case 3  & 2 & 2 & 0 & 2 & 2 & 4 \\
  Case 4  & 2 & 1 & 1 & 0 & 1 & 5 \\
  Case 5  & 1 & 1 & 1 & 1 & 0 & 6 \\
\bottomrule
\end{tabularx}
\end{table}

If the cluster number is set to three and two, all the possible ways to group the images are given in Tables \ref{tbcl3} and \ref{tbcl2}. Similarly, basing on the most indiscriminative visual representations of images gives the clustering of Case 1, which has the most indistinguishable pairwise comparisons. If we use the number of distinguishable pairwise comparisons as the quantity to measure the discriminativeness of the visual representations, these three tables are consistent: Case 1 to 4 in Table \ref{tbcl4} are same as Case 1 to 4 in Table \ref{tbcl3}, which have the same number of distinguishable pairwise comparisons and these numbers all increase from Case 1 to Case 4; Case 1 to 3 in Table \ref{tbcl3} are the same as Case 1 to 3 in Table \ref{tbcl2}, which have the same number of distinguishable pairwise comparisons and these numbers all increase from Case 1 to Case 3. Therefore, clearly, this quantity, i.e., the number of distinguishable pairwise comparisons, does not change with the setting of cluster number, which only depends on how to group images. The cluster number only affects the maximum number of the distinguishable pairwise comparisons. Noting that, under a certain rule, for example, k-means, the way to group images is completely determined by the representations; this quantity, in fact, is determined by the representations. Specifically, its value changes with the discriminativeness of the images' representations: the more discriminative are the representations, the larger is this quantity. Therefore, this quantity, the number of distinguishable pairwise comparisons, can be used to measure the discriminativeness of the images' representations through the model shown in Figure \ref{cl_rp}.
\begin{table}[H]
\caption{Five possible clusterings with cluster number as three. ``IND'': the number of indistinguishable pairwise comparisons; ``DIS'': the number of distinguishable pairwise comparisons. }
\label{tbcl3}
\setlength{\tabcolsep}{6mm}
\newcolumntype{B}{>{\centering\arraybackslash}X}
\begin{tabularx}{\textwidth}{llllll}
\toprule
 \textbf{Cluster} & \textbf{C1} & \textbf{C2} & \textbf{C3} & \textbf{Ind} & \textbf{Dis} \\
\midrule
  case 1  & 4 & 0 &  0 & 6 & 0 \\
  case 2  & 3 & 1 &  0 & 3 & 3 \\
  case 3  & 2 & 2 &  2 & 2 & 4 \\
  case 4  & 2 & 1 &  1 & 1 & 5 \\
\bottomrule
\end{tabularx}
\end{table}
\begin{table}[H]
\caption{Five possible clusterings with cluster number as 2. ``IND'': the number of indistinguishable pairwise comparisons; ``DIS'': the number of distinguishable pairwise comparisons. }
\label{tbcl2}
\setlength{\tabcolsep}{8mm}
\newcolumntype{B}{>{\centering\arraybackslash}X}
\begin{tabularx}{\textwidth}{lllll}
\toprule
 \textbf{Cluster} & \textbf{C1} & \textbf{C2} & \textbf{Ind} & \textbf{Dis} \\
\midrule
  case 1  & 4 & 0 & 6 & 0 \\
  case 2  & 3 & 1 & 3 & 3 \\
  case 3  & 2 & 2 & 2 & 4 \\
\bottomrule
\end{tabularx}
\end{table}

\subsection{Specific Toy Examples}

\begin{figure}[H]
\subfigure [c1]{
 \label{fig:c1}
\includegraphics[width=0.24\columnwidth]{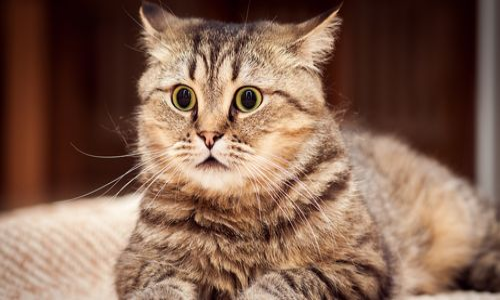}
}
\subfigure [c2] {
\label{fig:c2}
\includegraphics[width=0.24\columnwidth]{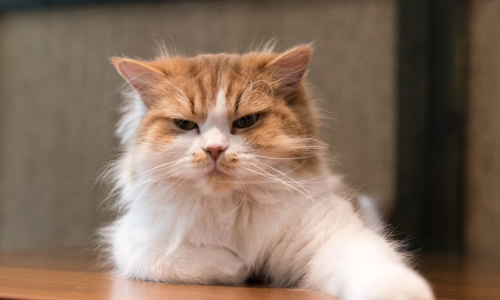}
}
\subfigure [c3] {
\label{fig:c3}
\includegraphics[width=0.24\columnwidth]{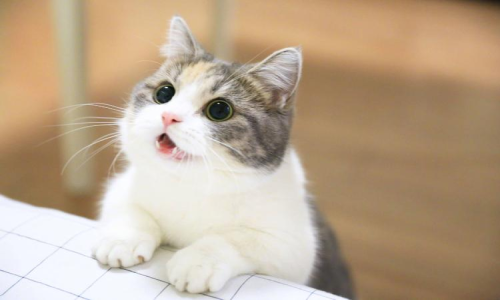}
}
\subfigure [d1]{
\label{fig:d1}
\includegraphics[width=0.24\columnwidth]{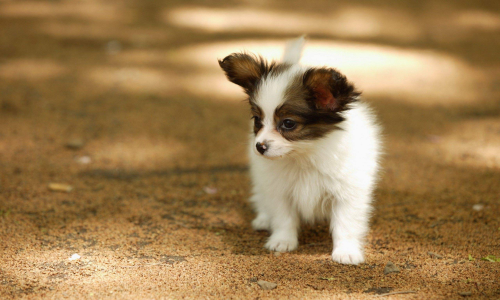}
}~~~~~~~~~~~~~~~~~~
\subfigure [d2]{
\label{fig:d2}
\includegraphics[width=0.24\columnwidth]{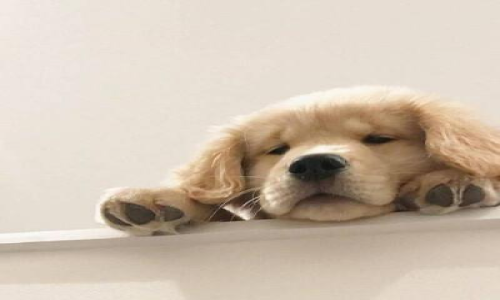}
}~~~~~~~~~~~~~~~~~~
\subfigure [d3]{
\label{fig:d3}
\includegraphics[width=0.24\columnwidth]{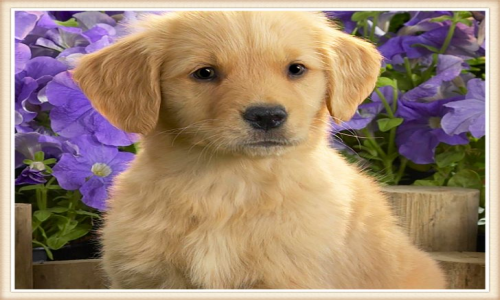}
}
\caption{Three cats: c1, c2, c3 (\textbf{a}--\textbf{c}); three dogs: d1, d2, d3 (\textbf{d}--\textbf{f}).}
\label{c-d}
\end{figure}

Let us consider the clustering of the images in Figure \ref{c-d} with the basic setting of the cluster number as three. Semantically, the best clustering is to group these images into two clusters: assigning the three cats c1, c2, c3 to one cluster while assigning the three dogs d1, d2, d3 to another cluster. However, according to {Remark \ref{md}}, to make representations of these images the most discriminative, we should group these images into three clusters, that is, assigning two images to each cluster. Assume that, based on the representations in two feature space $F_1$ and $F_2$, the six images are clustered through k-means as:
\begin{enumerate}
\item $F_1:$ cluster 1 $\{$c1, c2, c3$\}$; cluster 2 $\{$d1, d2, d3$\}$; cluster 3 $\{$ $\}$
\item $F_2:$ cluster 1 $\{$c1, c2$\}$; cluster 2 $\{$c3, d1$\}$; cluster 3 $\{$d2, d3$\}$
\end{enumerate}

Note that images are assigned into the same cluster because their representations are close, i.e., their representations are not distinguishable enough. On the other hand, in the representation learning model shown in Figure \ref{cl_rp}, images in the same cluster are assigned the same pseudo--labels, which will make the representations of the images in the same cluster closer. Therefore, we would assume that the representations of images in the same cluster as indistinguishable. Based on this assumption, we could list the pairs of distinguishable representations in $F_1$ and $F_2$ as:
\begin{enumerate}
\item $F_1:$ [c1, d1]; [c1, d2]; [c1, d3]; [c2, d1]; [c2, d2]; [c2, d3]; [c3, d1]; [c3, d2]; [c3, d3]
\item $F_2:$ [c1, c3]; [c1, d1]; [c1, d2]; [c1, d3]; [c2, c3]; [c2, d1]; [c2, d2]; [c2, d3]; [c3, d2]; [c3, d3]; [d1, d2]; [d1, d3]
\end{enumerate}

To compare the representations' degree of distinguishableness in $F_1$ and $F_2$, we neglect the same pairs in the lists above and have:
\begin{enumerate}
\item $F_1:$ [c3, d1]
\item $F_2:$ [c1, c3]; [c2, c3]; [d1, d2]; [d1, d3]
\end{enumerate}

These two lists indicate that representations of c3 and d1 are distinguishable in $F_1$ while are indistinguishable in $F_2$; the representations between c1 and c3, c2 and c3; and d1 and d2, d1 and d3 are distinguishable in $F_2$ while they are indistinguishable in $F_1$. From the viewpoints of the two lists above, assuming representations in $F_2$ are more discriminative than representations in $F_1$ is reasonable.  We have to emphasize that we are not performing semantic classification and that every image is different, although some of them belong to the same semantic class. The purpose of evenly distributed pseudo--labels is to make the representations of images as discriminative as possible when $k$ is fixed. Theoretically, as all the images are different, setting the cluster number is the same as the number of images (i.e., $k=N$) will make the representations of images the most discriminative. However, although discriminative representations is one of the most important targets in representation learning, it is not the only one. Good representations should also reveal the similarity of the images: close images have close representations; different images have very different representations. Therefore, in methods of instance discrimination, both ``positive'' and ``negative'' samples are very important. In performing instance discrimination, only the transformations of the same images are regarded as ``positive'', but in clustering-based methods, i.e., $k<N$, similar images are also regarded as ``positive'', which we usually cannot obtain only from the transformations of a image.   Therefore, setting $k=N$ may not be the best for visual-representation learning.

The example above  can only illustrate that grouping images into more clusters is better to find more discriminative representations. Let us consider another example: that of clustering the three cats c1, c2, c3 and one dog d1 into two clusters. Consider the representations in $F_1$ and $F_2$ as:
\begin{enumerate}
\item $F_1:$ cluster 1 $\{$c1, c2, c3$\}$; cluster 2 $\{$d1$\}$;
\item $F_2:$ cluster 1 $\{$c1, c2$\}$; cluster 2 $\{$c3, d1$\}$;
\end{enumerate}

Similarly, we can obtain different distinguishable pairwise comparisons in $F_1$ and $F_2$:
\begin{enumerate}
\item $F_1:$ [c3, d1]
\item $F_2:$ [c1, c3]; [c2, c3]
\end{enumerate}

Although the representations of c3 and d1 in $F_2$ are not distinguishable, the representations of c1 and c3, c2 and c3 are distinguishable. For this example, we can still reasonably assume that the representations of images in $F_2$ are more discriminative from the perspective of the two lists above. Again, every image is different; finding the representations that can reveal the differences amongst of images is one of the basic tasks in representation learning.

\section{Proof  {for Cases That $\bar n = N/k$ Is Not an Integer } } \label{appd}

For when $\bar n = N/k$ in (\ref{evcondn}) is not integer, the equalities (\ref{evcondn}) cannot be satisfied since all $n_i\;(i = 1, 2, \cdots, k)$ have to be integers. Setting $n_i =  \bar n + \hat n_i$, we can express (\ref{miniv}) as
\bea\label{ministd}
N_{\textrm{ind}} &=&\frac{1}{2} \big[(n_1^2+n_2^2+\cdots+n_k^2)-N\big]\nonumber \\
&=& \frac{1}{2} \big[\big((\bar n + \hat n_1)^2+(\bar n + \hat n_2)^2\nonumber
+\cdots+(\bar n + \hat n_k)^2\big)-N\big]\nonumber\\
&=& \frac{1}{2} \big[ \big( k \bar n^2 + 2\bar n (\hat n_1 + \hat n_2+\cdots+\hat n_k)\nonumber
+(\hat n_1^2 + \hat n_2^2+\cdots+\hat n_k^2)\big)-N \big].
\eea
Noting
\bq
k\bar n^2 = k \left(\frac{N}{k}\right)^2 = \frac{N^2}{k}
\eq
is a constant and
\bea
\hat n_1 + \hat n_2+\cdots+\hat n_k &=& (n_1 - \bar n) + (n_2 - \bar n)\nonumber
+ \cdots + (n_k - \bar n)\nonumber \\
 &=& n_1 + n_2 + \cdots + n_k - k \bar n \nonumber \\
 &=& 0
\eea
we obtain
\bea\label{ministd1}
N_{\textrm{ind}} &=&\frac{1}{2} \left[(\hat n_1^2+\hat n_2^2+\cdots+\hat n_k^2)+ \frac{N^2}{k} -N\right],
\eea
where $ (\hat n_1^2+\hat n_2^2+\cdots+\hat n_k^2) $ exactly is the square of standard deviation of distribution $\{n_1, n_2, \cdots, n_k\}$. Thus, $N_{\textrm{ind}}$ takes the minimal value when the standard deviation of the distribution $\{n_1, n_2, \cdots, n_k\}$ is the smallest, i.e., when the distribution $\{n_1, n_2, \cdots, n_k\}$ is the most uniform.

\section{Algorithm Analysis}\label{analysis}

Compared with similar work, e.g., self-labeling~\cite{conf/iclr/AsanoRV20a}, there are some differences in our method. First of all, the motivation of introducing evenly distributed pseudo--labels is different. In self-labeling~\cite{conf/iclr/AsanoRV20a}, even distribution is an assumption and introduced as a constraint to prevent models from degeneracy. In this paper, we create uniform pseudo--labels to make the images' representations as discriminative as possible. This is different because even distribution is not the only way to prevent degeneracy but to make representations as discriminative as possible, and evenly distributed, pseudo--labels are the best choice according to  { Remark \ref{md} } .

Secondly, the methods of approaching even distribution are very different. The technique used in~\cite{conf/iclr/AsanoRV20a} is from a classical algorithm in the optimal transport problem, the Sinkhorn--Knopp algorithm, while our method is based on the basic idea of frequency-sensitive competitive learning. In the Sinkhorn--Knopp algorithm, the manipulation of the output matrix $M_O$,
\bq\label{mo}
M_O =
\left(
  \begin{array}{cccc}
    o_1(1) & o_2(1) & \cdots & o_k(1)  \\
    o_1(2) & o_2(2) & \cdots & o_k(2)  \\
    \vdots & \vdots & \ddots & \vdots \\
    o_1(N) & o_2(N) & \cdots & o_k(N)  \\
  \end{array}
\right),
\eq
is multiplication while our algorithm only involves addition or subtraction, which makes our algorithm computationally much more efficient.

Last but not least, our method treats every output as a whole to make changes, that is, subtracting the same $k$-dimensional vector from every row (see (\ref{trans}), which is very important since it can preserve the neighborhood relation of any two outputs. To see that clearly, let us consider two arbitrary rows of $M_O$ (i.e., two arbitrary outputs), e.g., the $s_1$-th and $s_2$-th row. The difference in the two rows is
\bea
M_O[s_1,:]-M_O[s_2,:]=\big(o_1(s_1)-o_1(s_2), o_2(s_1)-o_2(s_2), \cdots, o_k(s_1)-o_k(s_2)\big),\nonumber
\eea
which will not change after translation (\ref{trans})
\bq\label{diff}
M'_O[s_1,:]-M'_O[s_2,:] = M_O[s_1,:]-M_O[s_2,:],
\eq
and their Euclidean distance will not change either
\bea\label{dist}
\big|M'_O[s_1,:]-M'_O[s_2,:]\big| &=& \big|M_O[s_1,:]-M_O[s_2,:]\big|\nonumber \\
&=&\big[\left(o_1(s_1)-o_1(s_2)\right)^2 
+ \left(o_2(s_1)-o_2(s_2)\right)^2 \nonumber \\
&&~~+\cdots + \left(o_k(s_1)-o_k(s_2)\right)^2\big]^{1/2}.
\eea

The equalities (\ref{diff}) and (\ref{dist}) indicate that iteratively performing translation (\ref{trans}) would not change the difference and Euclidean distance between any two output vectors. This invariance is consistent with smoothness (\ref{o'i}).

To clearly compare our algorithm with the Sinkhorn--Knopp algorithm used in~\cite{conf/iclr/AsanoRV20a}, we give the specific form of outputs after performing both algorithms:
\bq
M'_O(\textrm{ours}) =
\left(
  \begin{array}{cccc}
    o_1(1) - t_1 & o_2(1) - t_2 & \cdots & o_k(1) - t_k  \\
    o_1(2) - t_1 & o_2(2) - t_2 & \cdots & o_k(2) - t_k  \\
    \vdots & \vdots & \ddots & \vdots \\
    o_1(N) - t_1 & o_2(N) - t_2 & \cdots & o_k(N) - t_k  \\
  \end{array}
\right),
\eq
\bq\label{skagthm}
M'_O(\textrm{Sinkhorn-Knopp}) =
\left(
  \begin{array}{cccc}
   r_1 c_1 o_1(1) & r_1 c_2 o_2(1) & \cdots & r_1 c_k o_k(1)  \\
   r_2 c_1 o_1(2) & r_2 c_2 o_2(2) & \cdots & r_2 c_k o_k(2)  \\
    \vdots & \vdots & \ddots & \vdots \\
   r_k c_1 o_1(N) & r_k c_2 o_2(N) & \cdots & r_k c_k o_k(N)  \\
  \end{array}
\right).
\eq
where $r_1, \cdots, r_k$ and $c_1, \cdots, c_k$ are constants. Obviously, in general, making transformations as (\ref{skagthm})  cannot preserve equalities (\ref{diff}) and (\ref{dist}).

\section{Implementation Details }\label{im-details}

In the experiments conducted on AlexNet: 

All images were resized to $ 256\times 256 $ and then cropped to $ 224\times 224 $. Several augmentations were applied to inputs, such as color jitter, random grayscale, and so on. 
We trained all datasets with a batch size of 128 and a learning rate of 0.05 at the beginning. We trained CIFAR--10/100 for 1600 epochs in total and the pseudo--labels were updated around every two epochs. The learning rate dropped twice by multiplying 0.1 at epoch 960 and 1280. We trained SVHN for 400 epochs in total and the labels were updated nearly every epoch. The learning rate dropped twice by multiplying 0.1 at epoch 240 and 320.  We trained ImageNet for 450 epochs in total and the labels were updated nearly every epoch. The learning rate dropped three times by multiplying 0.1 at epoch 160, 300 and 380. As the type of images in CIFAR--10 and CIFAR--100 are very similar, we used the same augmentation strategy, such as random cropping, color jitter, horizontal flipping and so on. Regarding SVHN, we did not use the horizontal-flipping scheme since flip transformations are improper for digit learning. For the strong augmentations, we used the strategy of~\cite{conf/eccv/GansbekeVGPG20} with Augment = 8, n$\_$holes = 1, length = 16. We used SGD to optimize the network. 

In the experiments conducted on ResNet-50: 

All datasets kept their original size, and all the experiments used strong augmentations. For all experiments, $k$ was set to 128. The total training epochs was 300. Learning rate initally was 0.01 and then dropped to 0.001 and 0.0005 at epochs 160 and 240. Other parameters were the same as experiments on AlexNet.

\section{Unevenly Distributed pseudo--labels}\label{un-plbs}

To observe the effects of the distribution of the pseudo--labels on the performances of representation learning, we consider several unevenly distributed pseudo--labels. To do this, we only need to replace mean $\bar n(F)$ in (\ref{indicator}) with uneven distributions. In this paper we create the uneven distributions of target pseudo--labels as 
\bq
\tilde{n}_i(F) =\frac{N\;i^x}{\sum^k_i i^x},
\eq
where $N$ is the number of images and $k$ is the cluster number. For $x = 0, 2, 4, 6, 8, 10$, we give the distributions of pseudo--labels for CIFAR--10/100, i.e., $k=128$ for CIFAR--10 while $k=512$ for CIFAR--100, in Figure \ref{unp}.

\end{document}